\documentclass{article} 
\usepackage{iclr2024_conference,times}

\usepackage{amsmath,amsfonts,bm}

\def\eqref#1{equation~\ref{#1}}

\def\1{\bm{1}}

\DeclareMathAlphabet{\mathsfit}{\encodingdefault}{\sfdefault}{m}{sl}
\SetMathAlphabet{\mathsfit}{bold}{\encodingdefault}{\sfdefault}{bx}{n}

\usepackage{enumitem}
\usepackage{hyperref}
\usepackage{url}
\usepackage{color}
\usepackage{xcolor}
\usepackage{xfp,graphicx}
\usepackage{colortbl}
\usepackage{makecell}
\usepackage{supertabular}
\usepackage{array}
\usepackage{booktabs}

\definecolor{tabfirst}{rgb}{1, 0.7, 0.7}
\definecolor{tabsecond}{rgb}{1, 0.85, 0.7}
\definecolor{tabthird}{rgb}{1, 1, 0.7}

\title{GSFixer: Improving 3D Gaussian Splatting with Reference-Guided Video Diffusion Priors}

\author{Xingyilang Yin$^{1,2\dagger}$\thanks{Work done during an internship at VIVO. $\dagger$ Equal contribution. $\ddagger$ Corresponding authors.} \quad
\textbf{Qi Zhang$^{2\dagger}$} \quad 
\textbf{Jiahao Chang$^{3}$} \quad 
\textbf{Ying Feng$^{2}$} \\
\textbf{Qingnan Fan$^{2}$} \quad 
\textbf{Xi Yang$^{4}$} \quad 
\textbf{Chi-Man Pun$^{1\ddagger}$} \quad 
\textbf{Huaqi Zhang$^{2}$} \quad  
\textbf{Xiaodong Cun$^{5\ddagger}$} \\
$^{1}$University of Macau \ 
$^{2}$VIVO \
$^{3}$CUHKSZ \
$^{4}$Xidian University \
$^{5}$GVC Lab, Great Bay University \\
}

\usepackage{xspace}

\makeatletter
\DeclareRobustCommand\onedot{\futurelet\@let@token\@onedot}
\def\@onedot{\ifx\@let@token.\else.\null\fi\xspace}

\def\eg{\emph{e.g}\onedot}

\makeatother

\iclrfinalcopy 
\begin{document}

\maketitle

\begin{figure}[h]
  \centering
   \includegraphics[width=0.95\linewidth]{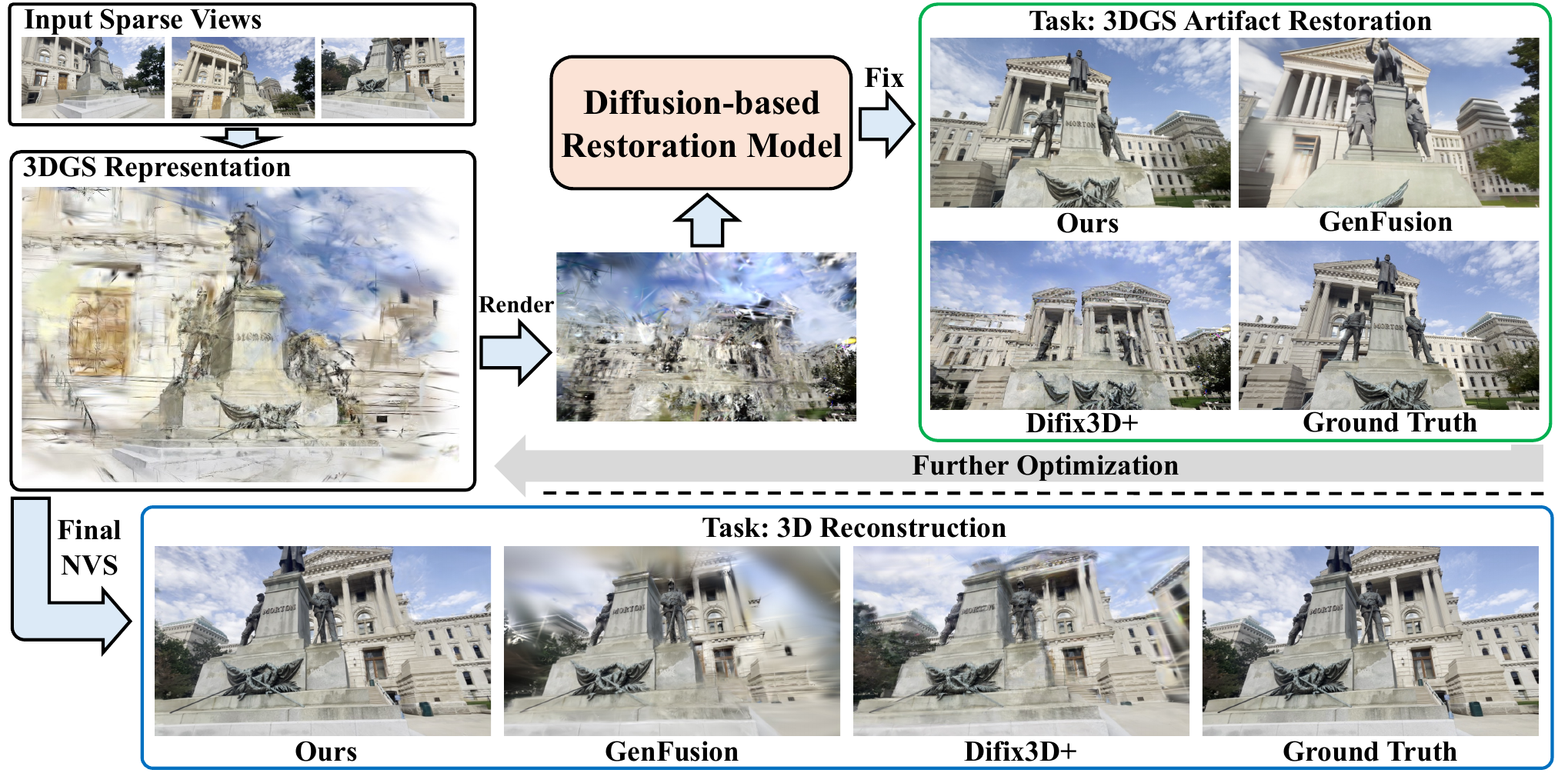}
   \caption{We introduce GSFixer, a framework capable of improving 3DGS in both artifact restoration (top) and 3D reconstruction (bottom) under sparse-view settings. Recent generative methods struggle with maintaining consistency between generated and input views. GSFixer guides the video diffusion model conditioned on both 3D and 2D signals to enhance consistency in novel view restoration, thereby improving 3D reconstruction quality.}
   \label{fig1: teaser}
\end{figure}

\begin{abstract}
Reconstructing 3D scenes using 3D Gaussian Splatting (3DGS) from sparse views is an ill-posed problem due to insufficient information, often resulting in noticeable artifacts. While recent approaches have sought to leverage generative priors to complete information for under-constrained regions, they struggle to generate content that remains consistent with input observations. To address this challenge, we propose \textbf{GSFixer}, a novel framework designed to improve the quality of 3DGS representations reconstructed from sparse inputs. The core of our approach is the reference-guided video restoration model, built upon a DiT-based video diffusion model trained on paired artifact 3DGS renders and clean frames with additional reference-based conditions. Considering the input sparse views as references, our model integrates both 2D semantic features and 3D geometric features of reference views extracted from the visual geometry foundation model, enhancing the semantic coherence and 3D consistency when fixing artifact novel views. Furthermore, considering the lack of suitable benchmarks for 3DGS artifact restoration evaluation, we present DL3DV-Res which contains artifact frames rendered using low-quality 3DGS. Extensive experiments demonstrate our GSFixer outperforms current state-of-the-art methods in 3DGS artifact restoration and sparse-view 3D reconstruction. Project page: \url{https://github.com/GVCLab/GSFixer}.

\end{abstract}

\section{Introduction}
3D reconstruction and novel view synthesis (NVS) are fundamental tasks in computer vision and graphics, with wide-ranging real-world applications in virtual reality, autonomous driving, and robotics. Recently, 3D Gaussian Splatting (3DGS)~\citep{kerbl20233d} has achieved impressive results in both reconstruction quality and rendering efficiency when dense input views are available. However, its performance degrades significantly in sparse-view settings, where limited viewpoint information leads to under-constrained 3D representations. In such cases, 3DGS often suffers from severe artifacts, including distorted geometric structures and incomplete reconstructions, particularly in less-observed regions or extreme novel viewpoints. These limitations hinder its applicability in real-world scenarios where acquiring dense multi-view data is challenging.

To alleviate the limitations, some previous regularization methods have been proposed to introduce additional constraints into the 3DGS optimization process, such as monocular depth~\citep{li2024dngaussian, zhu2024fsgs}, frequency smoothness~\citep{zhang2024fregs}, and random dropout~\citep{xu2025dropoutgs}. While these approaches can help prevent 3DGS representations from overfitting to sparse input views, they often remain sensitive to noise and yield only marginal improvements in NVS rendering quality. Inspired by the success of ReconFusion~\citep{wu2024reconfusion}, which introduces diffusion model into NeRF~\citep{mildenhall2020nerf} optimization, more recent studies~\citep{liu20243dgs, liu2024reconx, wu2025difix3d+, wu2025genfusion} explore incorporating 3DGS optimization with powerful generative priors from diffusion models, which are trained on internet-scale data. These strong priors enable the correction of spurious geometry or the inpainting of plausible content in novel views. However, a key challenge still remains: maintaining visual and 3D consistency between the generated and original input images, especially when the novel views are far from the observed inputs.

Meanwhile, recent advances in controllable video generation have demonstrated the effectiveness of incorporating various conditional signals~\citep{niu2024mofa, hu2024animate, he2025cameractrl, cui2025hallo3, yu2025trajectorycrafter} to guide video diffusion models in generating high-quality and coherent content. This progress motivates us to investigate appropriate control conditions to guide the video diffusion model in restoring artifact-ridden novel views, aiming for enhanced visual quality and 3D consistency. Given that the input to the sparse-view reconstruction task consists of only a few images—treated as ground-truth views—it is natural to align the restored novel views with these inputs. Thus, we exploit the information from reference views within the input set to condition the video diffusion model. Since artifacts manifest in 2D image space, their correction should rely on 2D visual priors to ensure semantic consistency with the reference views. At the same time, these artifacts originate from inaccurate 3DGS representations in 3D space, suggesting that effective restoration should also consider 3D structure. Therefore, an effective controllable video diffusion model for this task needs to integrate both 2D and 3D priors from reference views.

Inspired by the above motivation, we propose \textbf{GSFixer}, a novel generative reconstruction framework built upon a DiT-based video diffusion model~\citep{yang2024cogvideox}, trained on paired artifact 3DGS renders and clean frames, conditioned on additional reference-based conditions. Our key insight is to leverage information of reference views to guide the video diffusion model to restore artifact-prone novel views in a geometrically consistent and visually faithful manner. This enables high-quality novel view restoration and 3D reconstruction, as shown in Figure~\ref{fig1: teaser}. Specifically, we extract tokens via visual encoders (\eg, DINOv2~\citep{oquab2024dinov2}) as 2D semantic signal to ensure semantic consistency between the fixed novel views and the input observations. To further enforce multi-view consistency, we incorporate features from feed-forward 3D reconstruction networks (\eg, VGGT~\citep{wang2025vggt}) as 3D geometric condition. Moreover, we introduce a reference trajectory sampling strategy into the iterative generative optimization process to fix artifacts in novel views, effectively balancing angular coverage and restoration quality. In addition, to facilitate the evaluation of artifact removal capabilities of generative methods, we present DL3DV-Res, a benchmark comprising artifact-ridden frames rendered from low-quality 3DGS representations. Empirical results demonstrate that our GSFixer achieves superior sparse-view restoration and reconstruction performance on various challenging scenes.

Our main contributions are summarized as follows:
\begin{itemize}[leftmargin=*]
    \item We introduce GSFixer, a novel generative reconstruction framework tailored for improving the quality of 3DGS representations. It integrates a reference-guided video restoration model along with a reference-guided trajectory sampling strategy.
    \item We propose to incorporate video diffusion model conditioned on both 2D semantic and 3D geometric features extracted from the reference views. This effectively guides the restoration of artifact novel views, achieving both semantically coherent and 3D consistent results. 
    \item  We present DL3DV-Res benchmark for evaluating the performance of existing generative models in 3DGS artifact restoration. Extensive experiments demonstrate that our GSFixer outperforms existing baseslines in video-based 3D artifact restoration and sparse-view 3D reconstruction.
\end{itemize}

\section{Related Works}
\paragraph{Regularization Sparse-view Novel View Synthesis.}
Neural Radiance Fields (NeRF) and 3DGS have revolutionized the field of NVS by leveraging advances in neural rendering. While NeRF and 3DGS achieve remarkable photorealistic novel-view synthesis results given dense input views, they often suffer from severe overfitting issues and significant artifacts when training with sparse views. Previous works have attempted to address this limitation by introducing additional constraints and regularization into the per-scene 3D optimize process, such as depth supervision~\citep{deng2022depth, roessle2022dense, wang2023sparsenerf, li2024dngaussian, zhu2024fsgs}, normal consistency~\citep{yu2022monosdf, seo2023flipnerf}, smoothness priors~\citep{niemeyer2022regnerf, yang2023freenerf, zhang2024fregs}, semantic coherence~\citep{jain2021putting, truong2023sparf}, and random dropout strategy~\citep{xu2025dropoutgs}. However, these approaches often yield marginal improvements and remain sensitive to some specific scene data.

\paragraph{Conditional Video Diffusion Models.} Recently, generative models~\citep{rombach2022high, liu2023zero, blattmann2023stable, chen2023videocrafter1, chen2024videocrafter2, xing2024dynamicrafter, yang2024cogvideox} have advanced rapidly and demonstrate impressive capabilities in high-quality visual content generation. Building on this progress, recent efforts have extended video diffusion models with various control signals, such as depth~\citep{hu2025depthcrafter, zhao2024stereocrafter}, trajectory~\citep{niu2024mofa, yu2025trajectorycrafter}, audio~\citep{cui2025hallo3, kong2025let}, camera parameters~\citep{wang2024motionctrl, zhang2025recapture}, and point cloud renderings~\citep{yu2024viewcrafter, gu2025diffusion}. The additional control signals enable the diffusion models to generate coherent content with the user's inputs. Our work builds upon a video diffusion framework that leverages 3DGS renderings and features from clean reference frames as condition signals to guide the restoration of artifacts in novel views.

\paragraph{Generative Sparse-view Novel View Synthesis.} Generative models have demonstrated strong ability to inpaint plausible content in unobserved regions and restore degraded areas with high visual fidelity. Recent approaches~\citep{wu2024reconfusion, liu20243dgs, liu2024reconx, wu2025difix3d+, bao2025free360, wu2025genfusion} have made notable progress in the sparse-view NVS task by leveraging priors from generative models. For instance, ReconFusion~\citep{wu2024reconfusion} combines image diffusion with PixelNeRF~\citep{yu2021pixelnerf} as condition to optimize NeRF representations. Similiarly, more recent works~\citep{liu20243dgs, wu2025difix3d+, bao2025free360, wu2025genfusion} utilize diffusion priors to enhance low-qualtiy 3DGS representations. Our work is closely related to 3DGS-Enhancer~\citep{liu20243dgs}, GenFusion~\citep{wu2025genfusion} and DIFIX3D+~\citep{wu2025difix3d+}, all of which finetune diffusion models to correct artifacts in novel views. While following a similar direction, our method differs in two key aspects: (i) we finetune a DiT-based video diffusion model controlled with additional reference view conditions, and (ii) we incorporate both 2D semantic features and 3D geometric features extracted from visual geometry foundation models to guide the video restoration process, effectively fixing the artifacts in novel views.

\section{Methods}

\begin{figure}[t]
  \centering
   \includegraphics[width=0.95\linewidth]{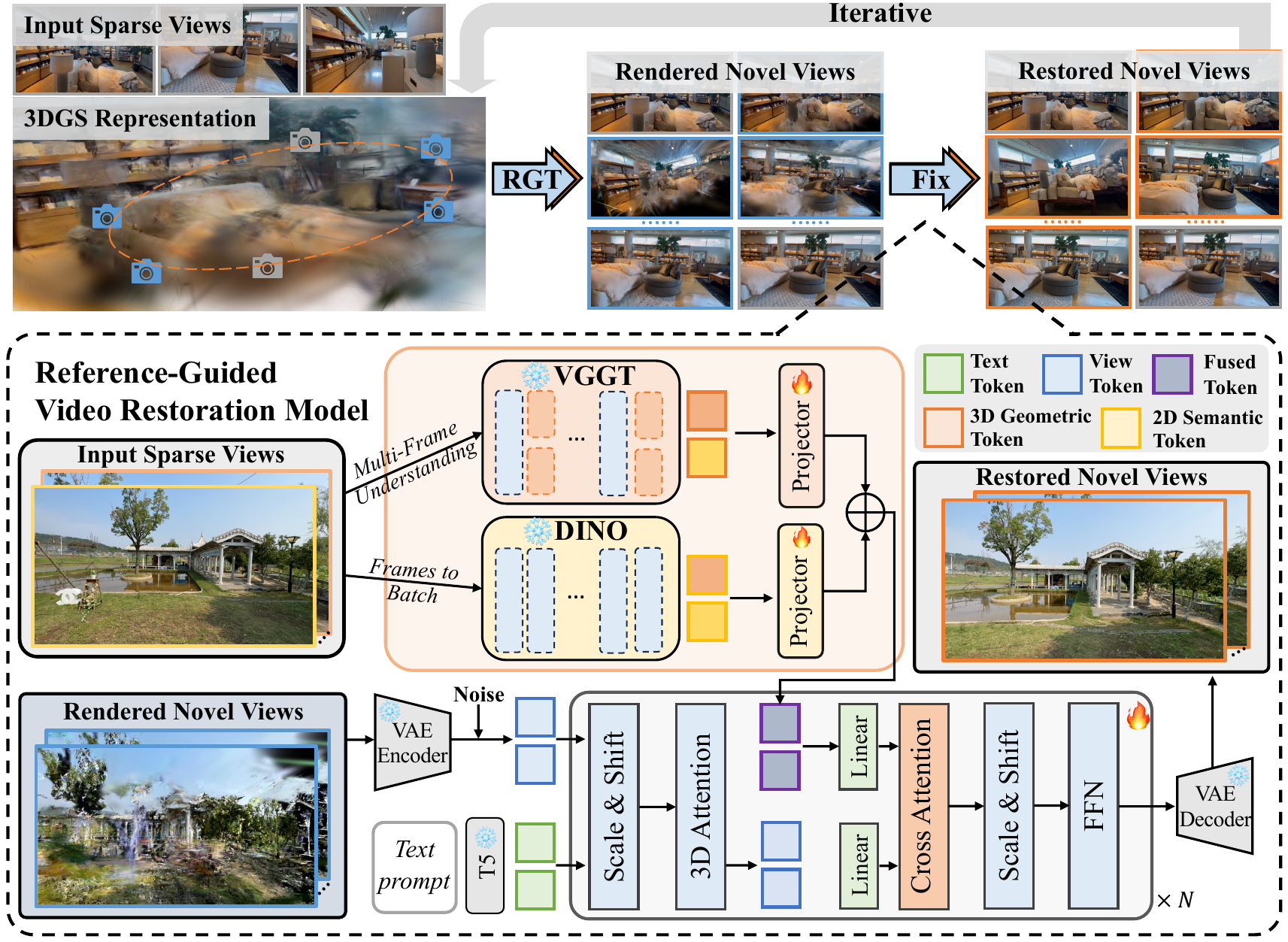}
   \caption{\textbf{Pipeline of GSFixer.} Given sparse-view images and their corresponding low-quality 3DGS representation, we render artifact-prone novel views between two reference views along a reference-guided trajectory. These novel views are fed into reference-guided video restoration model to correct artifacts, and the fixed novel views are then distilled back into the 3DGS representation to improve its quality. The restoration network is finetuned from CogVideoX and trained on paired artifact-ridden 3DGS renders and ground truth frames. It is additionally conditioned on 3D geometric tokens and 2D semantic tokens extracted from the reference views using pretrained VGGT and DINOv2 encoder, respectively.}
   \label{fig: GSFixer framework}
\end{figure}

\subsection{Preliminary}
\textbf{3D Gaussian Splatting and Novel View Synthesis.} 3DGS~\citep{kerbl20233d} explicitly represents a 3D scene as a collection of 3D Gaussian spheres, enabling high-quality 3d reconstructions and efficient novel view synthesis. Each 3D Gaussian sphere is defined by its center location $\mu$, scaling vector $s$, rotation quaternion $q$, opacity $\sigma$, and spherical harmonic (SH) coefficients $sh$. Thus, the Gaussian distribution is formulated as:
\begin{equation}
G(x)=e^{-\frac{1}{2}(x-\mu)^{T}\Sigma^{-1}(x-\mu)},
\end{equation}
where $\Sigma=RSS^{T}R^{T}$, $S$ denotes the scaling matrix corresponding to $s$ and $R$ is the rotation matrix determined by $q$. For rendering novel views, volume rendering integrates these elements using:
\begin{equation}
C = \sum_{i \in M} c_i \alpha_i \prod_{j=1}^{i-1} (1 - \alpha_j),
\end{equation}
where $C$ represents the final pixel color, which is computed via alpha blending of the view-dependent colors $c_i$ of the $M$ contributing Gaussians, weighted by their opacities $\alpha_i$..

\textbf{Video Diffusion Model.} Video diffusion models~\citep{blattmann2023stable, chen2023videocrafter1, chen2024videocrafter2, xing2024dynamicrafter} typically consist of two key stages: a forward diffusion process, which progressively inject noise $\epsilon$ into clean video data $\bm{x}_0 \in \mathbb{R}^{\text{n}\times \text{3}\times \text{h}\times \text{w}}$, yielding noisy samples $\bm{x}_t = \alpha_t\bm{x}_0 + \sigma_t \epsilon$ at each time step $t$;
and a reverse denoising process $p_\theta$, where a noise predictor $\epsilon\theta$ learns to recover the original data by removing the noise. This predictor was traditionally implemented using a U-Net architecture and is trained to minimize the following denoising objective:
\begin{equation}
\min_{\theta} \mathbb{E}_{t\sim\mathcal{U}(0,1),\epsilon\sim\mathcal{N}(\bm{0},\bm{I})}[\|\epsilon_\theta(\bm{x}_t,t) - \epsilon\|_2^2].
\end{equation}
Following Sora~\citep{brooks2024video}, recent approaches~\citep{yang2024cogvideox, wan2025wan} adopt the Diffusion Transformer (DiT)~\citep{peebles2023scalable} architecture for the noise predictor. During training, a pretrained 3D VAE encoder $\mathcal{E}$ compresses videos into latent space $\bm{z}=\mathcal{E}(\bm{x})$. These latent tokens $\bm{z}$ are then patchified, concatenated with text tokens, and fed into the DiT. At inference time, the model progressively denoises the latent tokens into clean tokens, which are subsequently decoded by the 3D VAE decoder $\mathcal{D}$ to generate the final video $\hat{\bm{x}}=\mathcal{D}(\bm{z})$.

\subsection{Overview of GSFixer}
The illustration of our GSFixer framework is depicted in Figure~\ref{fig: GSFixer framework}. Given $K$ sparse-view RGB images $\{I_i\}_{i=1,...,K}$, camera poses $\{P_i\}_{i=1,...,K}$, and corresponding trained 3DGS representation of the scene, our goal is to fix the artifacts present in novel views rendered by the low-quality 3DGS representation. We fix the artifact of novel views using our reference-guided video restoration model and distill the fixed images back into the 3DGS representation to improve its quality. In our reference-guided video restoration model, we employ a visual encoder $E_v$ and a spatial encoder $E_s$ to respectively extract 2D semantic tokens $T_{2D}$ and 3D geometry tokens $T_{3D}$ from the reference views $\{I_{r}^{ref}\}_{r=1,...,K}$ (Sec. \ref{sec: Reference-based Conditions}). These tokens are subsequently projected, fused and injected into a video diffusion process to guide the restoration of artifact-ridden novel views, preserving both semantic fidelity and 3D consistency with the input views (Sec.~\ref{sec: Reference-Guided Generation}). Moreover, we introduce a reference-guided trajectory sampling strategy within the iterative generative optimization to further enhance the quality. (Sec.~\ref{sec: Reference-Guided Generative Reconstruction}).

\subsection{Reference-guided Video Restoration Model}
\label{sec: Reference-guided Video Restoration Model}
\paragraph{Motivation.} Recent works~\citep{liu20243dgs, wu2025genfusion} have introduced video diffusion priors to restore artifact-prone novel views into clean frames. However, the generated frames often lack visual and 3D consistency with the input sparse views, leading to suboptimal 3D reconstruction performance. To address this, considering the artifacts finally lie in the 2D image space and are caused by suboptimal 3DGS representations in 3D space, we propose injecting both 2D semantic and 3D geometric control signals of reference views to guide the video diffusion process, enabling semantic and 3D consistency in restorating the artifact novel views. 


As illustrated in Figure~\ref{fig: GSFixer framework}, our reference-guided video restoration model is built upon DiT-based video diffusion model CogVideoX~\citep{yang2024cogvideox}, an image-to-video diffusion model capable of animating an input image to generate a video. We utilize its 3D Variational Autoencoder (VAE) for video compression and decompression. Additionally, we use BLIP~\citep{li2022blip} and T5~\citep{raffel2020exploring} encoder to caption video and extract the text tokens $t_{i}^{1D}$.

In our video-to-video restoration task, which aims to restore artifact frames into clean ones, we replace the original image condition with the 3DGS renders $\{I_{n}^{nov}\}_{n=1,\ldots,N}$ between two reference views $\{I_i, I_j\}$. Both the artifact frames and reference views are encoded using the 3D VAE encoder $\mathcal{E}$, concatenated with per-frame initial noise, and projected into view tokens $t^{view}$. These view tokens are then combined with text tokens $t^{text}$ and fed into the DiT blocks for denoising.

\label{sec: Reference-based Conditions}
\paragraph{Reference-based Conditions.} Given a set of sparse input images, we select two images $\{I_i, I_j\}$ as reference views, a set of novel views $\{I_{n}^{nov}\}_{n=1,\ldots,N}$ along arbitrary trajectories between them, we aim to learn a conditional $\bm{x} \sim p(\bm{x} | I^{nov}, \{I_i, I_j\})$ to guide the video diffusion model. To incorporate 2D semantic guidance into the video diffusion process, we employ a pretrained DINOv2~\citep{oquab2024dinov2} model as a 2D visual tokenizer. For each reference view $I_i$, DINOv2 encoder $\mathcal{E}_{2D}$ divides it into $L$ patches and extracts robust 2D visual features as a sequence of tokens:
\begin{equation}
t_{i}^{2D} = \mathcal{E}_{2D}(I_i), t_{i}^{2D} \in \mathbb{R}^{L \times C},
\end{equation}
where $t_{i}^{2D}$ denotes the resulting 2D semantic tokens, with $L$ and $C$ representing the number of tokens and the feature dimension, respectively.

To further control our video diffusion with 3D geometric priors, we adopt the pretrained Visual Geometry Grounded Transformer (VGGT)~\citep{wang2025vggt} as 3D geometric tokenizer. VGGT employs a 3D geometric encoder $\mathcal{E}_{3D}$, which composes several frame-wise and global self-attention layers, to encode multi-view reference views $\{I_i, I_j\}$:
\begin{equation}
t_{i}^{3D}, t_{j}^{3D} = \mathcal{E}_{3D}(I_i, I_j), t_{i}^{3D}\in \mathbb{R}^{L \times C}, t_{j}^{3D} \in \mathbb{R}^{L \times C}, 
\end{equation}
where $t_{i}^{3D}, t_{j}^{3D}$ represents the extracted 3D geometric tokens, with $L$ and $C$ representing the number of tokens and the feature dimension, respectively. VGGT leverages these tokens to produce all key 3D attributes of a scene, including camera parameters, depth maps, point maps, and 3D point tracks through various prediction heads. Consequently, the 3D geometric tokens include rich and robust 3D geometric priors of the scene.

\label{sec: Reference-Guided Generation}
\paragraph{Reference-guided Generation.}
As previously mentioned, the fixed frames may still lack visual and 3D consistency with the reference images. To address this, we fuse 3D geometric tokens and 2D semantic tokens of reference views to fusion tokens as additional condition:
\begin{equation}
t_{i}^{fusion} = Projector_{3D}(t_{i}^{3D}) + Projector_{2D}(t_{i}^{2D}),
\end{equation}
where $Projector_{3D}$ and $Projector_{2D}$ are implemented using linear and normalization layers.

To inject the reference fusion tokens to control the video diffusion process, we augment each DiT block by adding cross-attention layer after the 3D attention layer. In this cross-attention  mechanism, the view tokens serve as queries, while the fusion tokens act as keys and values:
\begin{equation}
t^{view}=CrossAttention(t^{view}, t^{fusion}).
\end{equation}
This allows the rich 2D and 3D information from the reference fusion tokens to be effectively injected into the view tokens, enabling direct alignment with the reference views and enhancing both semantic and geometric consistency when restoring artifact-prone novel views to clean ones.

\begin{figure}[t]
  \centering
   \includegraphics[width=0.90\linewidth]{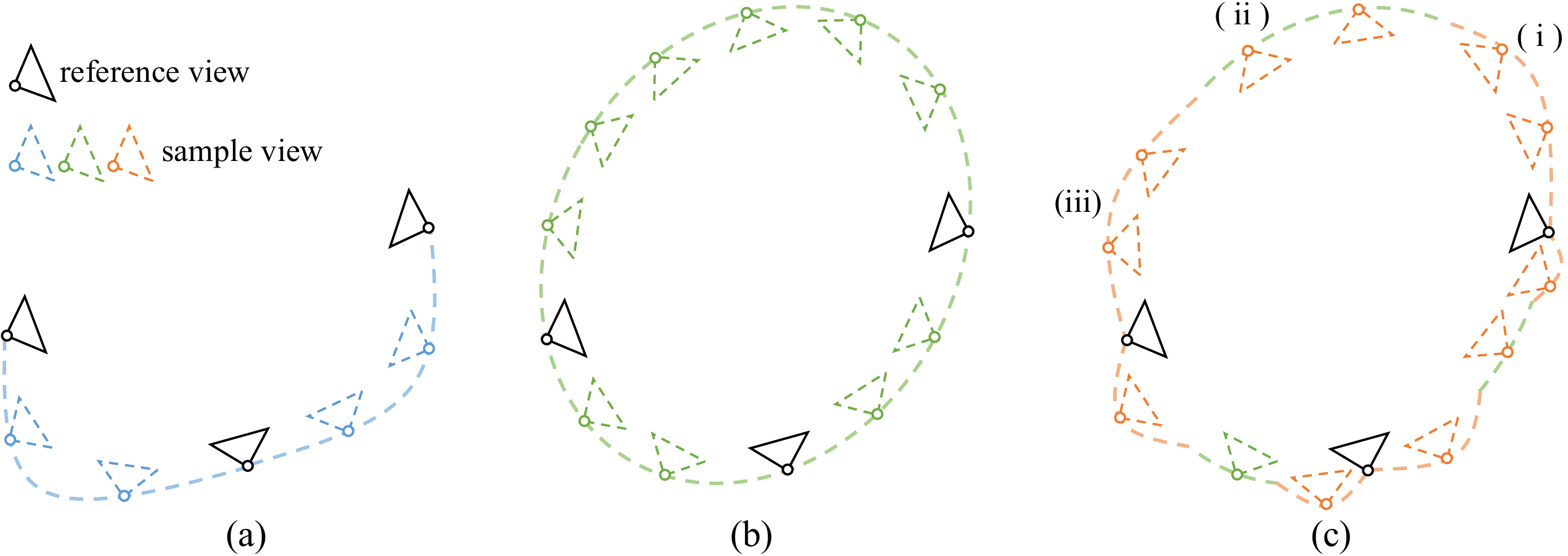}
   \caption{\textbf{Illustration of different trajectories.} (a) Interpolation trajectory: blue curve. (b) Ellipse trajectory: green curve. (c) Reference-guided trajectory: orange and green curve.}
   \label{fig: Reference-guided Trajectory}
\end{figure}

\subsection{Reference-Guided Generative Reconstruction}
\label{sec: Reference-Guided Generative Reconstruction}
Given rendered artifact-prone novel views and reference views, our trained reference-guided restoration diffusion model can produce consistent, artifact-free frames aligned with the input sparse views. Leveraging this capability, we supervise 3DGS optimization using both the input sparse views and the fixed novel views in an iterative training manner~\cite{haque2023instruct, wu2025genfusion, wu2025difix3d+}. Specifically, we begin by constructing an initial low-quality 3DGS representation from the sparse views. We then render novel views along new camera trajectories and feed these artifact-prone frames into our reference-guided video restoration model to obtain artifact-free outputs, as illustrated in Figure~\ref{fig: GSFixer framework}. These fixed novel views are subsequently added to the training set to further supervise the 3DGS optimization in a iterative way. 

\paragraph{Reference-guided Trajectory.} Trajectory sampling is critical in this iterative optimization process. Common sampling strategies, such as interpolation trajectory between input poses (Figure~\ref{fig: Reference-guided Trajectory} (a)) or ellipse trajectory along a spherical path across all the camera poses (Figure~\ref{fig: Reference-guided Trajectory} (b)) , have limitations. For our GSFixer, The interpolation trajectory yields high-quality fixed novel views but lacks angular diversity, while the ellipse trajectory provides broader view coverage but results in suboptimal fixed views. To balance these trade-offs, we propose a reference-guided trajectory sampling strategy, as shown in Figure~\ref{fig: Reference-guided Trajectory} (c). Specifically, (i) we first interpolate from a reference view to its nearest viewpoint on the spherical path, (ii) sample additional views along the sphere path, (iii) and then interpolate to the next nearest reference view. This hybrid sampling strategy achieves high-quality fixed views and angle coverage, leading to better 3DGS representation.

During optimization, we freeze the video diffusion model and supervise the 3DGS representation using a loss function $\mathcal{L}$ composed of two components: the reconstruction loss $\mathcal{L}_{recon}$ between rendered images and the input sparse views, and the generative loss $\mathcal{L}_{gen}$ between the rendered novel views and the corresponding fixed novel views. Specifically, we adopt simple photometric losses:
\begin{equation}
\mathcal{L}_{recon} = \lambda_{l1} \cdot \mathcal{L}_{l1} + \lambda_{SSIM} \cdot \mathcal{L}_{SSIM}.
\end{equation}
The generative loss $\mathcal{L}_{gen}$ consists of the same components as $\mathcal{L}_{recon}$ but is applied to the rendered and fixed novel views:
\begin{equation}
\mathcal{L} = \mathcal{L}_{recon} + \lambda \cdot \mathcal{L}_{gen},
\end{equation}
where we employ an annealing strategy~\cite{wu2025genfusion} to gradually increase the weight of the generative loss $\lambda$ during training.

\begin{table}[t]
\centering
\resizebox{0.95\linewidth}{!}{
\begin{tabular}{c|c|c|c|c|c|c|c|c}
    & PSNR $\uparrow$ &SSIM $\uparrow$ & LPIPS $\downarrow$ & I2V\_SC $\uparrow$ & I2V\_BC $\uparrow$ & OC $\uparrow$ & TF $\uparrow$ & MS $\uparrow$ \\
    \hline
    
Artifact~\citep{kerbl20233d} & 14.12 & 0.405 & 0.509 & 0.9382 & 0.9524 & 0.2066 &0.9058 &0.9548 \\

Difix3D+~\citep{wu2025difix3d+}  & 14.14  & 0.419  & 0.455 & 0.9288 & 0.9481 & 0.2393 & 0.9038 &0.9473 \\
GenFusion~\citep{wu2025genfusion}  & 14.56  & 0.453  &  0.486 & 0.8916 & 0.9258 & 0.2372 & 0.9135 & 0.9596\\
GSFixer (Ours)  & \textbf{16.72}  & \textbf{0.520} &  \textbf{0.399}  &  \textbf{0.9553}  &  \textbf{0.9644} & \textbf{0.2407} & \textbf{0.9233} & \textbf{ 0.9665}

    \end{tabular}
}
\caption{\textbf{Quantitative comparison on DL3DV-Res Dataset.} We compare the video 3D artifact restoration results with baselines. The best results are highlighted in \textbf{bold}.}
\label{tab:com_DL3DV_Res}
\end{table}

\begin{figure}[t]
  \centering
   \includegraphics[width=0.98\linewidth]{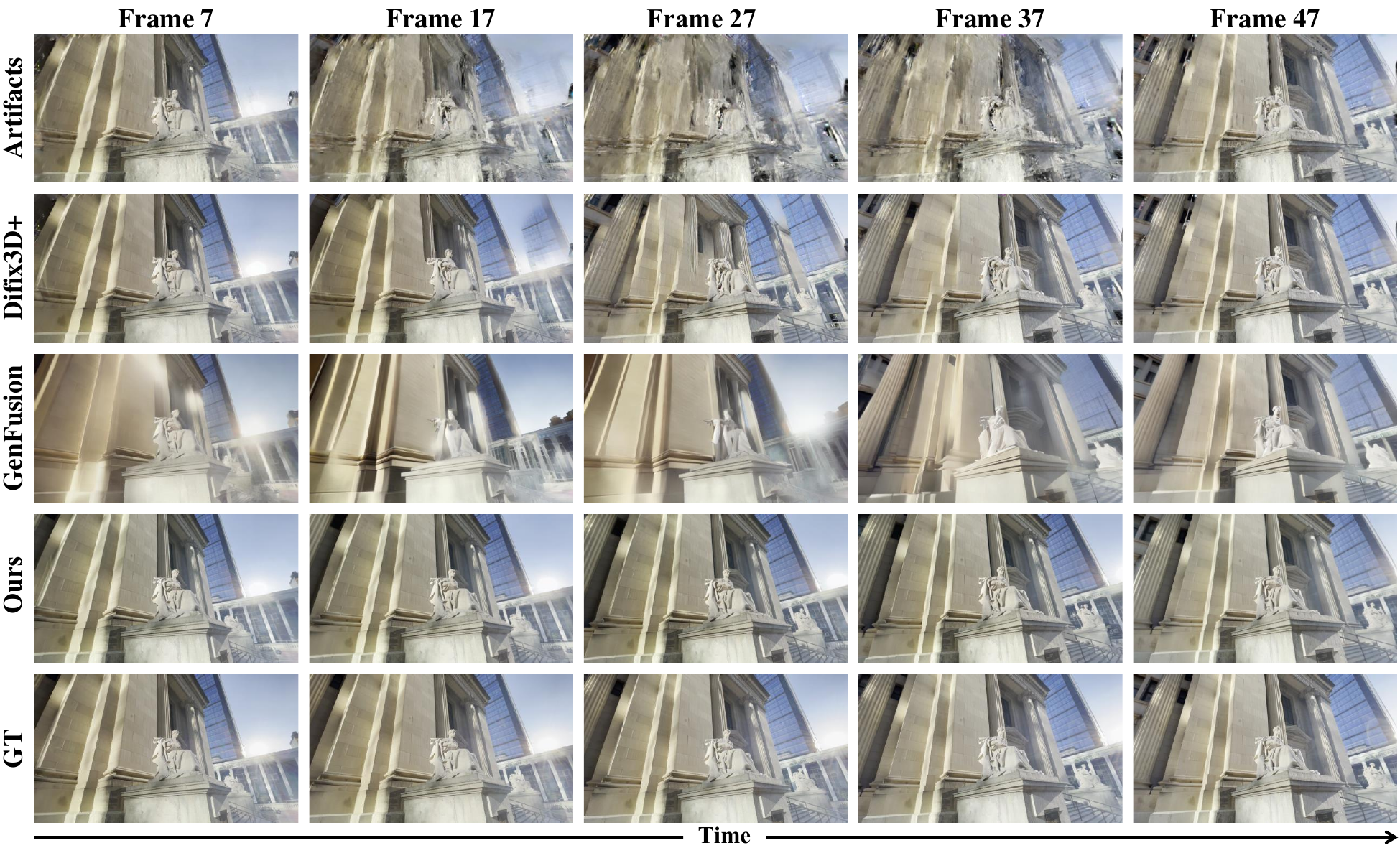}
   \caption{\textbf{Qualitative comparison on DL3DV-Res Benchmark}. We compare 3DGS artifact restoration quality of the existing generative methods.}
   \label{fig: 3DGS Artifact Restoration}
\end{figure}

\section{Experiments}
\subsection{Experimental Setup}
\textbf{Training Dataset and Evaluation Benchmarks.} We train our GSFixer on a random selection of 1,000 scenes from the DL3DV-10K dataset~\citep{ling2024dl3dv}. To construct paired artifact-laden and clean 3DGS renders, we employ a sparse-view 3D reconstruction strategy. Specifically, we use a few of input views (e.g., 3, 6, or 9) to reconstruct low-quality 3DGS representations. We then select camera trajectories from the dataset that include both the input views (used as reference views) and novel viewpoints (serving as ground truth frames). By rendering the low-quality 3DGS along these trajectories, we can generate the corresponding artifact-containing frames. Additionally, we extract the 1D text captions, 2D semantic tokens and 3D geometric tokens of the reference views by the pretrained BLIP, DINOv2 and VGGT encoders, respectively.

We evaluate our method and other baselines on 137 scenes from our proposed DL3DV-Res benchmark for the 3DGS artifact restoration task, and 28 scenes from the DL3DV-Benchmark~\citep{ling2024dl3dv} and 9 scenes from the Mip-NeRF 360 dataset~\citep{barron2022mip} for the sparse-view 3D reconstruction task. For evaluation, we report PSNR, SSIM, and LPIPS metrics and Overall Consistency (OC), Temporal Flickering (TF), Motion Smoothness (MS) scores of VBench~\citep{huang2024vbench} and I2V\_Subject Consistency (I2V\_SC), I2V\_Background Consistency (I2V\_BC) scores of VBench++~\citep{huang2024vbench++} protocol.

The DL3DV-Res benchmark is constructed from all available scenes in the DL3DV-Benchmark, using a similar sparse-view reconstruction protocol. We train 3DGS models with extremely sparse input views (e.g., 3) and render the reconstructions along the original camera trajectories of each scene. This produces novel views with severe artifacts, paired with high-quality ground truth images. For fair comparison, all the methods initialize with the COLMAP~\citep{schonberger2016structure} point cloud in all the experiments and we filter and retain only the visible points from the input sparse training views. 

\begin{table}[t]
\centering
\resizebox{0.95\linewidth}{!}{
\begin{tabular}{@{}l@{\,\,}|ccc|ccc|ccc}
    & \multicolumn{3}{c|}{PSNR $\uparrow$} & \multicolumn{3}{c|}{SSIM $\uparrow$} & \multicolumn{3}{c}{LPIPS $\downarrow$}  \\
    & 3-view & 6-view & 9-view  & 3-view & 6-view & 9-view  & 3-view & 6-view & 9-view   \\ \hline
    
3DGS~\citep{kerbl20233d} & 13.72 & 17.11 & 19.05 & 0.410 & 0.547 &  0.625  & 0.521  & 0.372 & 0.293  \\
Difix3D+~\citep{wu2025difix3d+}   & \cellcolor{tabsecond}15.07 & \cellcolor{tabthird}18.26 & \cellcolor{tabthird}20.12 & \cellcolor{tabthird}0.481  & \cellcolor{tabthird}0.589 & \cellcolor{tabthird}0.656  & \cellcolor{tabfirst}0.473 & \cellcolor{tabfirst}0.329  & \cellcolor{tabfirst}0.259  \\
GenFusion~\citep{wu2025genfusion}  & \cellcolor{tabthird}14.64  & \cellcolor{tabsecond}18.36 &  \cellcolor{tabsecond}20.32 & \cellcolor{tabsecond}0.498  & \cellcolor{tabsecond}0.610 & \cellcolor{tabfirst}0.688  & \cellcolor{tabthird}0.493 & \cellcolor{tabthird}0.374  &\cellcolor{tabthird}0.314 \\
GSFixer (Ours)  & \cellcolor{tabfirst}16.21 & \cellcolor{tabfirst}19.11 &\cellcolor{tabfirst}20.60  &\cellcolor{tabfirst}0.536  & \cellcolor{tabfirst}0.625  & \cellcolor{tabsecond}0.675   & \cellcolor{tabsecond}0.478 & \cellcolor{tabsecond}0.360 & \cellcolor{tabsecond} 0.307

    \end{tabular}
}
\caption{\textbf{Quantitative comparison on DL3DV-Benchmark Dataset.} We compare the rendering quality with baselines given 3, 6 and 9 views.}
\label{tab:com_sparse_view_recon_dl3dv}
\end{table}

\begin{figure}[t]
  \centering
   \includegraphics[width=0.98\linewidth]{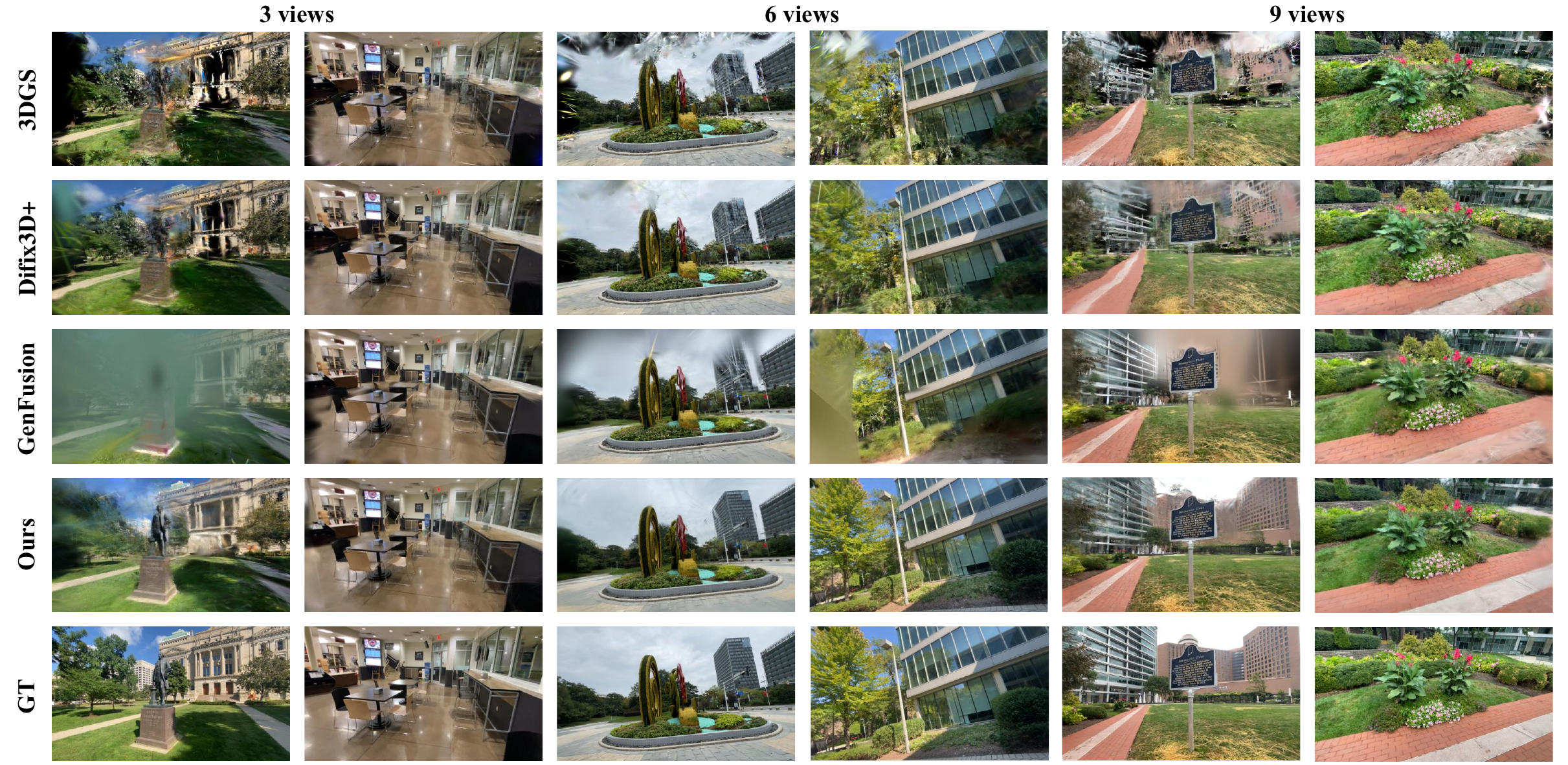}
   \caption{\textbf{Qualitative comparison on DL3DV-Benchmark.} We compare the novel view with baselines rendering quality using 3, 6, and 9 input views.}
   \label{fig4: Qualitative comparison on DL3DV-Benchmark}
\end{figure}

\textbf{Implementation Details.} We implement and initialize the parameters of our reference-guided video diffusion model based on the pretrained CogVideoX-5B-I2V~\citep{yang2024cogvideox}. During training, the frame resolution is fixed at $480{\times}720$, and the video length is set to 49 frames. The training stage is conducted for 10,000 iterations with a learning rate of $2{\times}10^{-5}$, incorporating a
warm-up strategy, and optimized using the AdamW optimizer. The proposed model is trained on 8 NVIDIA H20 GPUs with a batch size of 8 for about 4 days.

\subsection{Comparison with Other Methods}
\textbf{3DGS Artifact Restoration.} To evaluate the effectiveness of existing models in 3DGS artifact restoration, we compare our GSFixer with recent generative methods, including Difix3D+~\citep{wu2025difix3d+} and GenFusion~\citep{wu2025genfusion} on our proposed DL3DV-Res benchmark. As shown in Table~\ref{tab:com_DL3DV_Res}, the quantitative results demonstrate that GSFixer significantly outperforms both Difix3D+ and GenFusion in correcting artifacts in novel views across all pixel-wise metrics. Specifically, GSFixer achieves improvements of 2.16 in PSNR, 0.067 in SSIM, and 0.087 in LIPIS compared to GenFusion. Furthermore, our method consistently outperforms the baselines across all VBench and VBench++ metrics, showcasing state-of-the-art performance in 3DGS artifact restoration. It is noteworthy that the I2V\_SC and I2V\_BC scores of Difix3D+ and GenFusion are even lower than the original artifact frames after their artifact correction, suggesting that these methods struggle to maintain consistency when fixing artifacts in novel views. The qualitative comparisons in Figure~\ref{fig: 3DGS Artifact Restoration} further emphasize GSFixer's superior quality and consistency in fixing the artifacts of novel views. For instance, Difix3D+ and GenFusion fail to restore the content consistently, such as the foreground statue and background buildings. See supplement for more details.

\begin{table}[t]
\centering
\resizebox{0.95\linewidth}{!}{
\begin{tabular}{@{}l@{\,\,}|ccc|ccc|ccc}
    & \multicolumn{3}{c|}{PSNR $\uparrow$} & \multicolumn{3}{c|}{SSIM $\uparrow$} & \multicolumn{3}{c}{LPIPS $\downarrow$}  \\
    & 3-view & 6-view & 9-view  & 3-view & 6-view & 9-view  & 3-view & 6-view & 9-view   \\ \hline
    
Zip-NeRF$^\dag$~\citep{barron2023zip} & 12.77 & 13.61 & 14.30 & 0.271 & 0.284 & 0.312  & 0.705 & 0.663 & 0.633  \\
FreeNeRF$^\dag$~\citep{yang2023freenerf} & 12.87 & 13.35 & 14.59 & 0.260 & 0.283 & 0.319  & 0.715 & 0.717 & 0.695  \\
SimpleNeRF$^\dag$~\citep{somraj2023simplenerf} & 13.27 & 13.67 & 15.15 & 0.283 & 0.312 & 0.354  & 0.741 & 0.721 & 0.676  \\
ZeroNVS$^\dag$~\citep{sargent2024zeronvs} &  14.44 & 15.51 & 15.99  & 0.316 & 0.337 & 0.350  & 0.680 & 0.663 & 0.655  \\
ReconFusion$^\dag$~\citep{wu2024reconfusion} &  \cellcolor{tabsecond}15.50 &  \cellcolor{tabsecond}16.93 &  \cellcolor{tabthird}18.19  &  \cellcolor{tabsecond}0.358 &  0.401 & 0.432  & 0.585 & 0.544 & 0.511 \\
\hline
3DGS$^\dag$~\citep{kerbl20233d} & 13.06 & 14.96 & 16.79 & 0.251 & 0.355 &  0.447  &  \cellcolor{tabsecond}0.576 &  0.505 &  0.446  \\
2DGS$^\dag$~\citep{huang20242d} & 13.07 & 15.02 & 16.67  & 0.243 & 0.338 & 0.423  &  0.580 &  0.506 &  0.449  \\
FSGS$^\dag$~\citep{zhu2024fsgs} & 14.17 &  16.12 &  17.94  &  0.318 &  \cellcolor{tabthird}0.415 &  \cellcolor{tabfirst}0.492  &  \cellcolor{tabthird}0.578 & 0.517 & 0.468  \\

Difix3D+$^\ddag$~\citep{wu2025difix3d+}  &  13.92 &  15.94 &  17.54  &  0.298 &  0.382 &  0.452  & \cellcolor{tabthird}0.578 &  \cellcolor{tabfirst}0.468 &  \cellcolor{tabfirst}0.391 \\

GenFusion$^\ddag$~\citep{wu2025genfusion}  &  \cellcolor{tabthird}15.03 &  \cellcolor{tabthird}16.90 &  \cellcolor{tabsecond}18.29  &  \cellcolor{tabthird}0.357 &  \cellcolor{tabfirst}0.430 &  \cellcolor{tabsecond}0.489  & \cellcolor{tabthird}0.578 &  \cellcolor{tabthird}0.494 &  \cellcolor{tabthird}0.440 \\

GSFixer (Ours)  & \cellcolor{tabfirst}15.61  & \cellcolor{tabfirst}17.27  &  \cellcolor{tabfirst}18.63  & \cellcolor{tabfirst}0.370  & \cellcolor{tabsecond}0.426  & \cellcolor{tabthird}0.481   &\cellcolor{tabfirst}0.559 & \cellcolor{tabsecond}0.478  &\cellcolor{tabsecond}{ 0.420}

    \end{tabular}
}
\caption{\textbf{Quantitative Comparison on Mip-NeRF 360 Dataset.} We compare the rendering quality with baselines given 3, 6 and 9 views. $\dag$ denotes results reproduced by ReconFusion and GenFusion, while $\ddag$ indicates results reproduced by us on their official implementation.}
\label{tab:sparse_view_mipnerf}
\end{table}

\begin{figure}[t]
  \centering
   \includegraphics[width=0.98\linewidth]{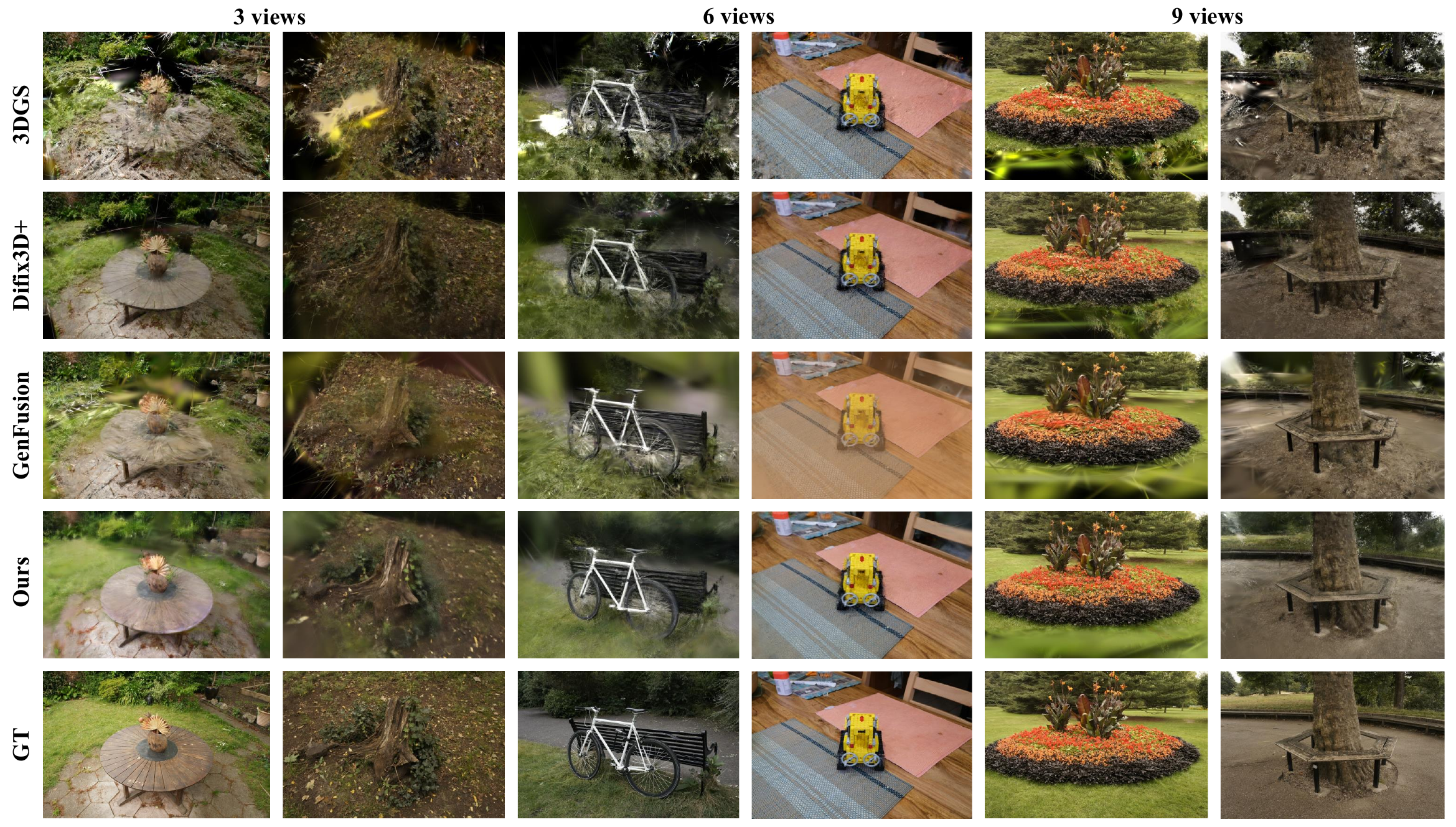}
   \caption{\textbf{Qualitative comparison on Mip-NeRF 360.} We compare the novel view rendering quality with baselines using 3, 6, and 9 input views.}
   \label{fig5: Qualitative comparison on Mip-NeRF 360}
\end{figure}

\textbf{In Domain Sparse View Reconstruction.} We next compare our method with baseline approaches for sparse-view 3D reconstruction on the DL3DV-Benchmark. As shown in Table~\ref{tab:com_sparse_view_recon_dl3dv} and Figure~\ref{fig4: Qualitative comparison on DL3DV-Benchmark}, our GSFixer achieves more realistic novel view synthesis across 3, 6, and 9 input views. For example, under the extremely sparse 3-view setting, GSFixer significantly improves 3DGS quality by 3.55 dB in PSNR, 0.119 in SSIM, and 0.034 in LPIPS. Moreover, our GSFixer outperforms state-of-the-art methods such as Difix3D+ and GenFusion by a substantial margin. Difix3D+ fails to generate plausible content in occluded or missing regions, as illustrated in the building structure in the first row of the figure. It also does not adequately fix geometry-distorted Gaussians, as seen in the road reconstruction in the fifth and sixth rows. Additionally, inconsistent novel views generated by Difix3D+'s image diffusion process degrade overall quality, particularly noticeable in the statue (first row) and in the building and sky regions (third row). Similarly, GenFusion exhibits notable artifacts in fixing 3DGS representation. For example, its video diffusion process generates ``foggy'' geometry in the background (statue in first row and building in fifth row), failing to produce 3D-consistent content in missing regions. This results in blurred renderings, as shown in the third and fourth rows. In contrast, our GSFixer effectively inpaints plausible content in missing regions and performs both semantically and geometrically consistent fixes in novel views, leading to high-quality and coherent novel view renderings.


\textbf{Out of Domain Sparse View Reconstruction.} To demonstrate the generalizability of our method for out-of-distribution dataset, we further evaluate the methods on the challenging Mip-NeRF 360 dataset with 3, 6, and 9 input views. The quantitative results presented in Table~\ref{tab:ablation_model_design_mipnerf360} show that our GSFixer outperforms the baseline approaches, highlighting its strong generalization performance on unseen and complex scenes. For instance, our GSFixer surpasses generative 3DGS-based methods such as Difix3D+ and GenFusion. Moreover, we observe that our GSFixer even outperforms generative NeRF-based ReconFusion in sparse view settings. This is particularly notable as 3DGS is generally more prone to overfitting on sparse input views compared to NeRF. Qualitative results shown in Figure~\ref{fig5: Qualitative comparison on Mip-NeRF 360} further illustrate the semantic and 3D consistent generative capability of our GSFixer. For detailed per-scene results, please refer to the supplementary material.

\begin{table}[t]
\centering
\resizebox{0.95\linewidth}{!}{
\begin{tabular}{c|c|c|c|c|c|c|c|c}
    & PSNR $\uparrow$ &SSIM $\uparrow$ & LPIPS $\downarrow$ & I2V\_SC $\uparrow$ & I2V\_BC $\uparrow$ & OC $\uparrow$ & TF $\uparrow$ & MS $\uparrow$ \\
    \hline
    Ours w/o 3D tokens & 16.36 & 0.510 & 0.414 & 0.9527 & 0.9630 & 0.2403 & 0.9218 & 0.9663\\
Ours w/o 2D tokens  & 16.48   & 0.516  & 0.409 & 0.9540 & 0.9635 & 0.2405 & 0.9225 & 0.9665 \\
Ours full model  & \textbf{16.72}  & \textbf{0.520} &  \textbf{0.399}  &  \textbf{0.9553}  &  \textbf{0.9644} & \textbf{0.2407} & \textbf{0.9233} & \textbf{ 0.9665}
    \end{tabular}
}
\caption{\textbf{Ablation study about different conditions.} We report the PSNR, SSIM, LPIPS, VBench and VBench++ metrics for the full model and its ablated versions on DL3DV-Res.}
\label{tab:ablation_model_design_dl3dv_res}
\end{table}

\begin{figure}[t]
  \centering
   \includegraphics[width=0.98\linewidth]{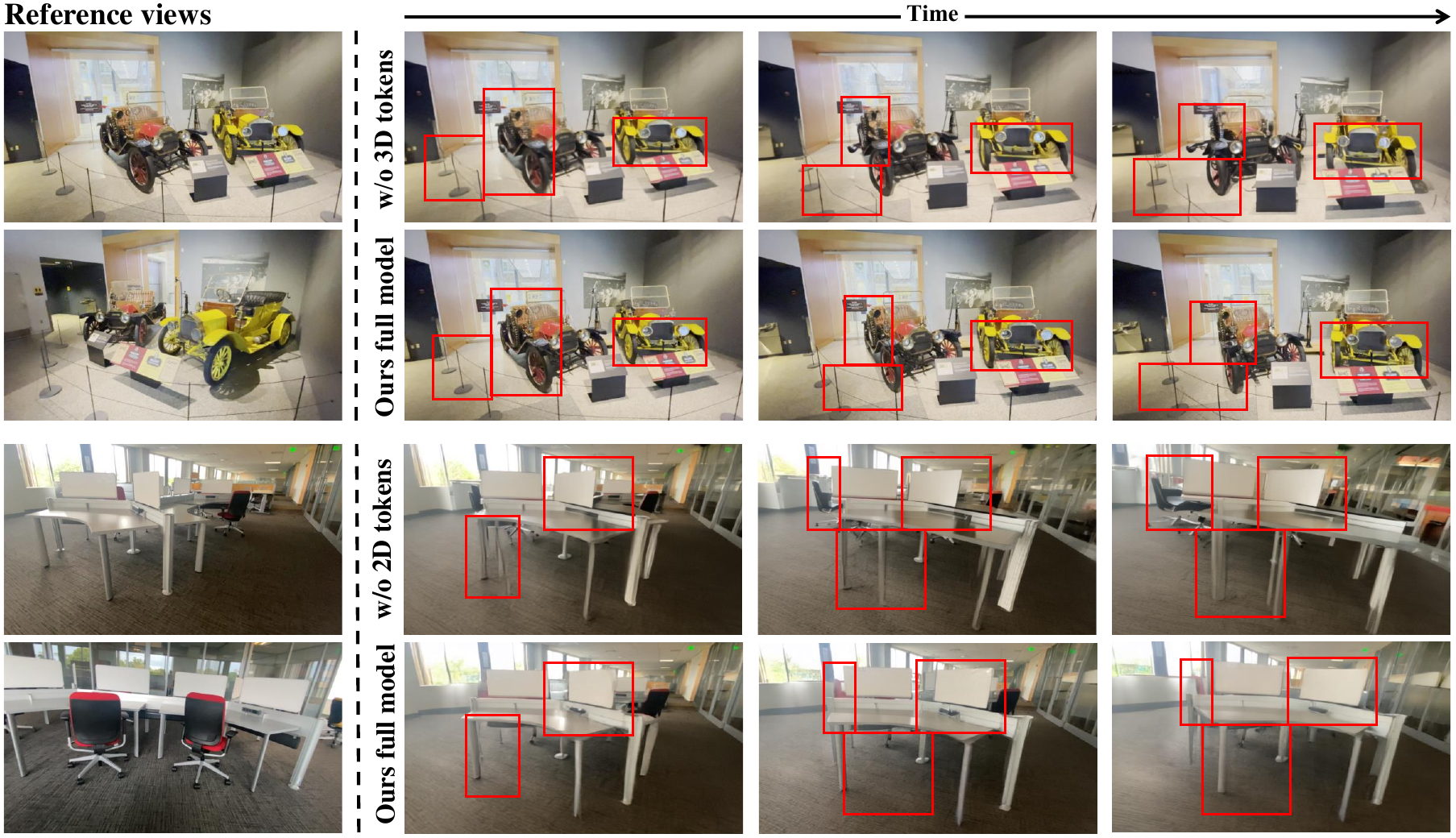}
   \caption{\textbf{Effectiveness of reference view conditions.} We compare our full model with two alternatives: a variant without 3D conditions (top), and a variant without 2D conditions (bottom). The red boxes highlight the most prominent differences.}
   \label{fig: ablation different conditions}
\end{figure}

\subsection{Ablation Studies}
\textbf{Effectiveness of Reference View Conditions.} We conduct an ablation study on one of our key contributions: the reference-guided video restoration model, which controls the 3DGS artifact restoration process using 3D and 2D tokens extracted from reference views. We compare the video restoration results of three variants on the DL3DV-Res benchmark: our full model, a version without 3D tokens, and one without 2D tokens. As quantitatively results presented in Table~\ref{tab:ablation_model_design_dl3dv_res}, our full model consistently outperforms the other two variants across all metrics. The qualitative comparisons in Figure~\ref{fig: ablation different conditions} further support this finding. From the visualization, we observe that removing the 3D condition leads to poor 3D consistency with the reference views, e.g., misalignment in the fence and cars (highlighted in red boxes). This is because 3DGS renders under sparse-view settings often exhibit severe geometric distortions, making it difficult for the diffusion model to generate geometrically consistent content without strong 3D priors. Similarly, without the 2D condition, the model fails to restore semantically plausible details, such as the table leg, baffle, socket, and chair marked in red boxes. In contrast, our full model successfully restores artifact-ridden novel views to frames that are both 3D-consistent and semantically aligned with the reference views, demonstrating the effectiveness of injecting both 3D and 2D conditions into the video diffusion model.

\begin{figure}[t]
  \centering
   \includegraphics[width=0.98\linewidth]{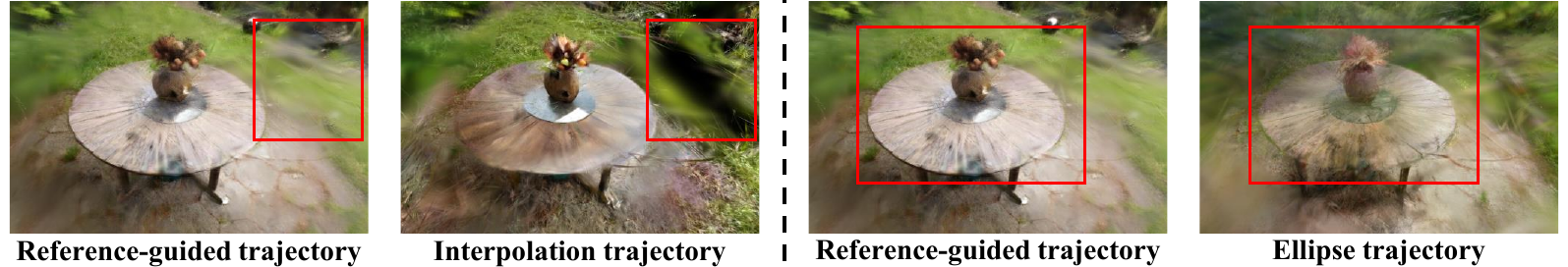}
   \caption{\textbf{Effectiveness of reference-guided trajectory.} We compare with widely used interpolation trajectory and ellipse trajectory. The red boxes highlight the most prominent differences.}
   \label{fig: ablation different trajectory}
\end{figure}

\begin{table}[t]
\centering
\begin{tabular}{c|c|c|c}
    & PSNR $\uparrow$ &SSIM $\uparrow$ & LPIPS $\downarrow$   \\
    \hline
    Interpolation trajectory & 14.50 & 0.353 & 0.565 \\
Ellipse trajectory & 15.46 & 0.362 & 0.563 \\
Reference-guided trajectory  & \textbf{15.61}  & \textbf{0.370} & \textbf{ 0.559} 

    \end{tabular}
\caption{\textbf{Ablation study about trajectory sampling.} We report the PSNR, SSIM, and LPIPS metrics for different trajetories on Mip-NeRF 360.}
\label{tab:ablation_model_design_mipnerf360}
\end{table}

\textbf{Effectiveness of Reference Guided Trajectory.} To validate the effectiveness of our reference guided-trajectory sampling strategy, we further conduct an ablation study on the challenging Mip-NeRF 360 dataset using a 3-view 3D reconstruction setting.  Qualitative results are presented in Figure~\ref{fig: ablation different trajectory}, with the most notable differences highlighted in red boxes. As shown in the figure, the interpolation trajectory lacks sufficient angular coverage, leading to incomplete reconstructions with missing regions. Meanwhile, the ellipse trajectory results in degraded rendering quality, as the novel views to be restored are not temporally sequential with the reference views. In contrast, our reference-guided trajectory ensures both coverage and quality in the reconstruction process. The quantitative results in Table~\ref{tab:ablation_model_design_mipnerf360} further confirm the effectiveness of the reference-guided trajectory sampling strategy.
\section{Conclusions}
We present GSFixer, a novel pipeline for enhancing the quality of 3D Gaussian Splatting in sparse-view 3D reconstruction and novel view synthesis. Our reference-guided video restoration model integrates both 3D geometric and 2D semantic condition signals from reference views, ensuring semantically coherent and geometrically consistent artifact correction in novel views. Furthermore, we introduce a reference-guided trajectory sampling strategy within the iterative reconstruction process, enabling high-quality 3D reconstruction. In addition, we propose the DL3DV-Res benchmark to evaluate the 3DGS artifact restoration capability. 
\paragraph{Limitations and Future Work.} Our GSFixer bulids upon a DiT-based video diffusion model that requires 50 denoising steps, which may limit the efficiency. As a 3D enhancement model, its performance is inherently limited by the quality of the initial 3DGS representation. Future work may explore improving GSFixer with advanced single-step video diffusion models and improved 3D representation, enabling efficient and high-fidelity novel view synthesis.

\bibliography{iclr2024_conference}
\bibliographystyle{iclr2024_conference}

\clearpage
\appendix
\section*{Appendix: GSFixer: Improving 3D Gaussian Splatting via Reference-Guided Video Diffusion Priors}


\section{Reference-Guided Video Diffusion Model Details}
Our reference-guided video diffusion model is built upon CogVideoX-5B-I2V, a pretrained DiT-based video diffusion model capable of generating 49-frame videos at a resolution 480$\times$720 from a single input image. In our video-to-video restoration task, which aims to restore artifact frames into clean ones, we replace the original image condition with the 3DGS renders $\{I_{n}^{nov}\}_{n=1,\ldots,47}$ between two reference views $\{I_1, I_2\}$. In the training process, each training sample consists of a sequence of 47-frame artifact-prone 3DGS renders, two reference images, and 49-frame ground truth RGB video. The artifact frames (49$\times$3$\times$480$\times$720) and ground truth frames (49$\times$3$\times$480$\times$720) are first  encoded into latent features using a 3DVAE, resulting in latent dimensions of 16$\times$13$\times$60$\times$90. The ground truth latent features are then perturbed with noise for the diffusion training process. For the view tokens, we concatenate, patchify and project the latent features to 17550$\times$3072 view tokens. For the text tokens, we use BLIP to generate text captions of the reference views and use T5 to extract 226$\times$4096 feature embedding, which are projected to 226$\times$3072 text tokens. To extract geometric and semantic conditioning signals, the two reference views are first resize to 350$\times$518 to feed into VGGT and DINOv2 encoder to produce 2$\times$930$\times$2048 geometric tokens and 2$\times$930$\times$1024 semantic tokens. We abondon 2$\times$5$\times$2048 camera token from VGGT, and remove 2$\times$1$\times$1024 class token and 2$\times$4$\times$1024 registration token from DINOv2, resulting in 2$\times$925$\times$2048 3D geometric tokens and 2$\times$925$\times$1024 2D semantic tokens. These are projected to 2$\times$925$\times$3072 using $Projector_{3D}$ and $Projector_{2D}$, respectively. $Projector_{3D}$ is a linear layer mapping from 2048 to 3072 dimensions, followed by LayerNorm. $Projector_{2D}$ is a linear layer mapping from 1024 to 3072 dimensions, followed by LayerNorm. Finally, the 3D geometric and 2D semantic tokens are combined and reshaped into fusion tokens of dimension 1850$\times$3072, which serve as rich 3D and 2D priors to guide the restoration process of diffusion model. During inference, we employ DDIM sampling with classifier-free guidance to modulate condition adherence strength.

\section{Details of Comparision Baselines}
We compare our method with (i) regularization methods and (ii) generative methods for sparse-view reconstruction and 3DGS artifact restoration tasks. Specifically, we evaluate NeRF-based per-scene regularization methods including Zip-NeRF~\cite{barron2023zip}, FreeNeRF~\cite{yang2023freenerf} and SimpleNeRF~\cite{somraj2023simplenerf}; NeRF-based generative methods such as ZeroNVS~\cite{sargent2024zeronvs} and ReconFusion~\cite{wu2024reconfusion}; 3DGS-based per-scene regularization method FSGS~\cite{zhu2024fsgs}; and 3DGS-based generative methods Difix3D+~\cite{wu2025difix3d+} and GenFusion~\cite{wu2025genfusion}. Difix3D+ incoporates generative priors from the image diffusion model SD-turbo~\cite{sauer2024adversarial}, finetuned on paired artifact renders and clean frames. In contrast, GenFusion bulids upon the video diffusion model DynamicCrafter~\cite{xing2024dynamicrafter} that condition video frames on artifact-prone RGBD renders.


\begin{figure}[t]
  \centering
   \includegraphics[width=0.92\linewidth]{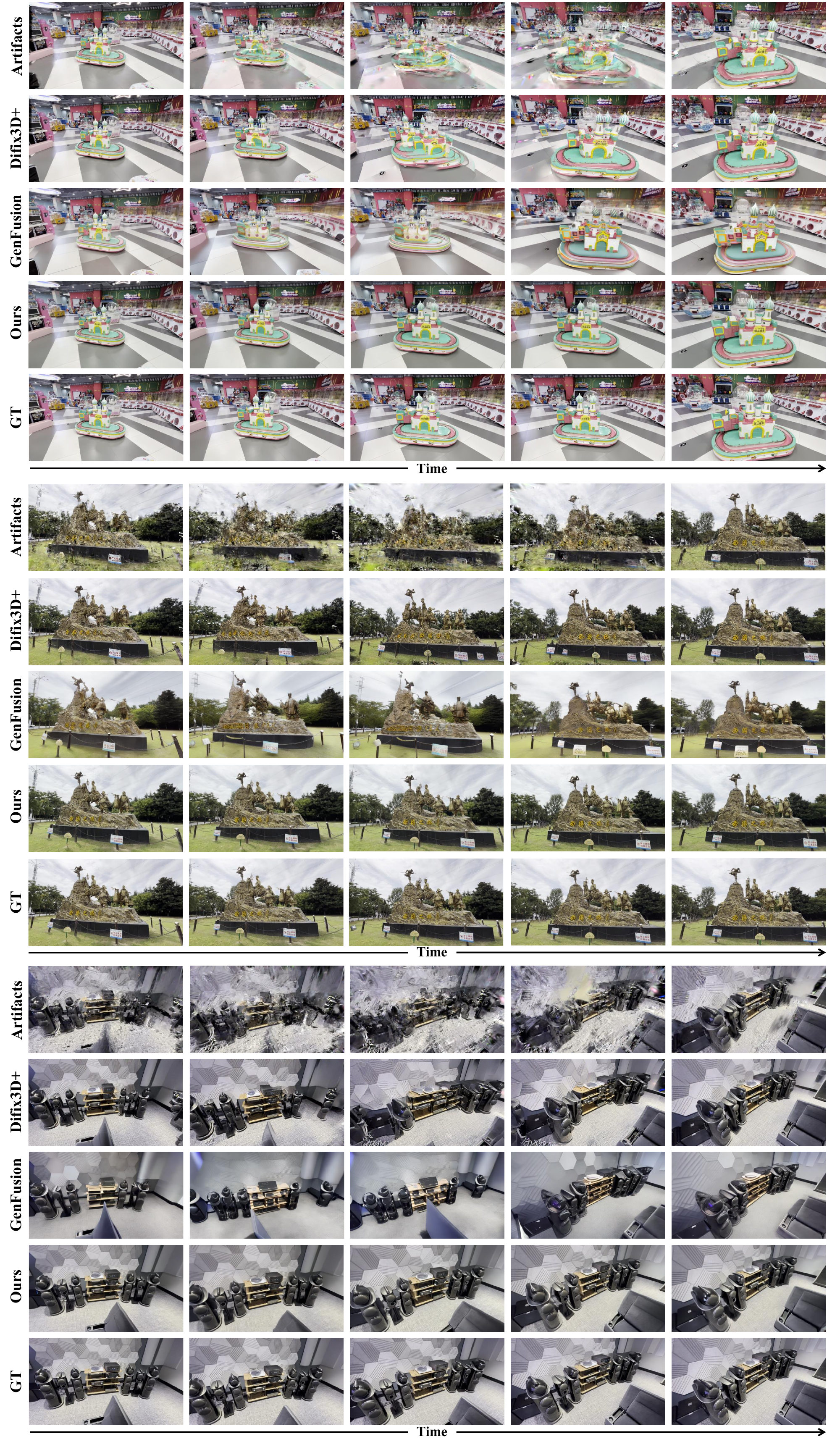}
   \caption{Qualitative comparision of 3DGS artifact restoration on DL3DV-Res.}
   \label{fig_sup: 3DGS Artifact Restoration}
\end{figure}

\section{More Quantitative and Qualitative Results}
We provide extensive per-scene experimental results in Figure~\ref{fig_sup: 3DGS Artifact Restoration}, Table~\ref{tab:dl3dv_Res_per_scene_video_restoration_part1}, Table~\ref{tab:dl3dv_Res_per_scene_video_restoration_part2}, Table~\ref{tab_sup:dl3dv_per_scene_3d_recon_sparse_view3}, Table~\ref{tab_sup:dl3dv_per_scene_3d_recon_sparse_view6}, Table~\ref{tab_sup:dl3dv_per_scene_3d_recon_sparse_view9}, and Table~\ref{tab_sup:mipnerf_per_scene_3d_recon}.

\begin{table}[ht]
    \renewcommand{\tabcolsep}{4.5pt}
    \centering
    \resizebox{0.98\linewidth}{!}{
    \begin{tabular}{l|ccc|ccc|ccc|ccc}
    & \multicolumn{3}{c|}{Artifact} & \multicolumn{3}{c|}{Difix3D+} & \multicolumn{3}{c|}{GenFusion}& \multicolumn{3}{c}{GSFixer (Ours)}\\
    & PSNR$\uparrow$ & SSIM$\uparrow$ & LPIPS$\downarrow$ & PSNR$\uparrow$ & SSIM$\uparrow$ & LPIPS$\downarrow$ & PSNR$\uparrow$ & SSIM$\uparrow$ & LPIPS$\downarrow$ & PSNR$\uparrow$ & SSIM$\uparrow$ & LPIPS$\downarrow$ \\
\hline
032dee9fb0 &17.43	&0.510	&0.447 &17.04	&0.521	&0.375 &17.03	&0.573	&0.412 &23.07	&0.709	&0.196\\
0569e83fdc &12.65	&0.218	&0.567 &12.42	&0.214	&0.536 &12.79	&0.250	&0.562 &15.32	&0.330	&0.458\\
06da796666 &13.09	&0.407	&0.525 &13.13	&0.429	&0.500 &14.76	&0.526	&0.507 &14.47	&0.513	&0.499\\
073f5a9b98 &11.50	&0.129	&0.613 &11.34	&0.121	&0.604 &12.19&	0.17&	0.626 &12.53	&0.181	&0.620\\
07d9f9724c &20.37	&0.680	&0.321 &20.06	&0.706	&0.269 &18.95	&0.662	&0.303 &25.59	&0.860	&0.123\\
0853979305 &17.24	&0.530	&0.442 &17.33	&0.549	&0.391 &17.55	&0.578	&0.419 &19.51	&0.625	&0.355\\
093ef327b4 &13.08	&0.445	&0.587 &12.85	&0.422	&0.540 &13.49	&0.452	&0.552 &14.96	&0.493	&0.472\\
0a1b7c20a9 &15.59	&0.507	&0.454 &15.27	&0.517	&0.403 &15.41	&0.529	&0.441 &19.11	&0.670	&0.273\\
0a485338bb &15.13	&0.467	&0.466 &15.24	&0.491	&0.402 &15.89	&0.491	&0.440 &20.23	&0.623	&0.290\\
0bfdd020cf &11.08	&0.397	&0.660 &11.23	&0.355	&0.560 &14.27	&0.432	&0.579 &14.87	&0.402	&0.486\\
119fd56d37 &12.41	&0.331	&0.599 &12.45	&0.360	&0.563 &13.13	&0.452	&0.563 &13.93	&0.468	&0.552\\
1264931635 &12.38	&0.205	&0.568 &12.32	&0.195	&0.538 &13.43	&0.245	&0.565 &14.86	&0.267	&0.536\\
14eb48a50e &12.13	&0.396	&0.554 &12.12	&0.428	&0.498 &12.77	&0.491	&0.521 &15.65	&0.558	&0.437\\
15ff83e253 &13.37	&0.360	&0.576 &13.59	&0.330	&0.491 &14.00	&0.389	&0.552 &15.65	&0.383	&0.474\\
165f5af8bf &14.59	&0.374	&0.492 &14.87	&0.396	&0.420 &15.47	&0.429	&0.462 &16.86	&0.477	&0.384\\
183dd248f6 &16.49	&0.522	&0.498 &16.36	&0.527	&0.453 &16.29	&0.559	&0.472 &16.80	&0.571	&0.458\\
1ba74c2267 &17.91	&0.606	&0.491 &18.43	&0.636	&0.403 &17.97	&0.666	&0.431 &20.34	&0.728	&0.345\\
1d6a9ed47c &12.21	&0.238	&0.512 &12.10	&0.238	&0.473 &12.74	&0.243	&0.533 &15.09	&0.342	&0.459\\
1da888bded &13.27	&0.385	&0.558 &13.30	&0.396	&0.501 &13.62	&0.434	&0.522 &16.08	&0.506	&0.417\\
1de58be515 &19.19	&0.707	&0.279 &20.09	&0.732	&0.197 &18.32	&0.629	&0.319 &24.18	&0.820	&0.165\\
2385549d39 &14.95	&0.388	&0.457 &15.30	&0.406	&0.364 &15.13	&0.378	&0.438 &18.42	&0.529	&0.292\\
26fd23358f &14.40	&0.353	&0.510 &14.19	&0.371	&0.448 &14.33	&0.385	&0.495 &17.60	&0.484	&0.363\\
286239bd0d &11.63	&0.249	&0.571 &11.57	&0.251	&0.552 &12.5	&0.304	&0.561 &13.15	&0.311	&0.546\\
2991a75d1f &12.54	&0.345	&0.609 &12.22	&0.313	&0.572 &12.52	&0.352	&0.615 &13.15	&0.355	&0.561\\
2b65ba886e &10.72	&0.232	&0.570 &10.57	&0.231	&0.557 &11.67	&0.272	&0.562 &12.27	&0.313	&0.540\\
2beaca3189 &12.30	&0.268	&0.580 &11.99	&0.271	&0.541 &12.74	&0.305	&0.570 &14.58	&0.373	&0.465\\
2cbfe28643 &16.33	&0.559	&0.425 &16.98	&0.606	&0.314 &16.49	&0.595	&0.395 &20.11	&0.725	&0.231\\
2f3e1c0f68 &13.57	&0.380	&0.519 &13.67	&0.409	&0.462 &14.28	&0.475	&0.482 &15.83	&0.511	&0.421\\
32c2b92fac &16.87	&0.610	&0.482 &16.75	&0.637	&0.401 &16.48	&0.662	&0.455 &20.56	&0.734	&0.324\\
341b4ff3df &15.35	&0.408	&0.478 &15.14	&0.414	&0.417 &15.18	&0.410	&0.457 &18.05	&0.558	&0.327\\
35317e6219 &12.02	&0.269	&0.586 &12.10	&0.283	&0.549 &13.86	&0.381	&0.529 &13.98	&0.388	&0.518\\
35872363e1 &14.94	&0.416	&0.478 &14.86	&0.428	&0.411 &15.29	&0.447	&0.430 &17.79	&0.534	&0.321\\
374ffd0c5f &18.72	&0.569	&0.418 &19.11	&0.604	&0.337 &17.99	&0.614	&0.394 &21.89	&0.723	&0.247\\
387eeb925b &13.14	&0.317	&0.551 &12.96	&0.332	&0.469 &13.39	&0.375	&0.528 &15.93	&0.461	&0.392\\
389a460ca1 &13.66	&0.523	&0.502 &13.58	&0.545	&0.435 &12.75	&0.556	&0.503 &15.08	&0.619	&0.432\\
3b16a10ec9 &12.67	&0.337	&0.516 &12.83	&0.357	&0.472 &13.92	&0.437	&0.487 &14.99	&0.460	&0.457\\
3b7529dccc &12.77	&0.316	&0.526 &12.57	&0.317	&0.505 &13.76	&0.367	&0.512 &14.12	&0.393	&0.480\\
3bb3bb4d3e &14.81	&0.561	&0.458 &14.92	&0.594	&0.393 &16.06	&0.658	&0.375 &18.12	&0.710	&0.297\\
3bb894d193 &15.94	&0.429	&0.426 &15.92	&0.449	&0.330 &16.57	&0.463	&0.380 &17.93	&0.573	&0.255\\
41036716da &14.98	&0.473	&0.502 &15.13	&0.508	&0.441 &15.00	&0.525	&0.466 &16.99	&0.593	&0.376\\
444da1b4a3 &12.16	&0.347	&0.515 &11.84	&0.365	&0.468 &11.85	&0.377	&0.486 &16.33	&0.564	&0.310\\
457e9a1ae7 &14.08	&0.434	&0.520 &13.90	&0.455	&0.474 &14.28	&0.495	&0.508 &16.21	&0.579	&0.425\\
484c0aca40 &13.87	&0.504	&0.492 &13.71	&0.536	&0.429 &13.86	&0.565	&0.472 &16.33	&0.663	&0.348\\
493816813d &12.17	&0.232	&0.563 &12.18	&0.207	&0.529 &13.65	&0.262	&0.537 &14.82	&0.282	&0.526\\
4ae797d07b &13.07	&0.378	&0.520 &12.80	&0.361	&0.491 &13.24	&0.386	&0.504 &15.26	&0.435	&0.449\\
4ff8650b5c &14.93	&0.470	&0.499 &14.84	&0.488	&0.457 &14.87	&0.518	&0.476 &16.16	&0.559	&0.404\\
50c46cf8b8 &15.72	&0.520	&0.502 &15.74	&0.554	&0.459 &15.75	&0.604	&0.478 &17.55	&0.640	&0.406\\
513e4ea2e8 &14.46	&0.417	&0.502 &14.62	&0.438	&0.424 &14.38	&0.455	&0.487 &20.25	&0.644	&0.227\\
54bf355ca7 &13.61	&0.380	&0.524 &13.51	&0.384	&0.496 &14.33	&0.441	&0.505&15.20	&0.459	&0.469\\
56452d9cd9 &14.39	&0.328	&0.517 &14.58	&0.345	&0.470 &15.28	&0.396	&0.501 &16.41	&0.421	&0.453\\
565553aa89 &13.51	&0.318	&0.527 &13.65	&0.338	&0.474 &14.67	&0.384	&0.487 &16.35	&0.453	&0.408\\
599ca3e04c &14.48	&0.453	&0.525 &14.22	&0.473	&0.477 &13.87	&0.478	&0.485 &15.53	&0.532	&0.448\\
5a27c00f52 &13.64	&0.384	&0.470 &13.40	&0.391	&0.400 &13.75	&0.395	&0.437 &17.23	&0.547	&0.286\\
5a69d1027b &10.68	&0.192	&0.661 &10.83	&0.168	&0.604 &13.23	&0.262	&0.629 &13.42	&0.240	&0.568\\
5c3af58102 &17.45	&0.563	&0.389 &17.94	&0.618	&0.280 &17.04	&0.592	&0.338 &22.53	&0.767	&0.180\\
5f0041e53d &18.02	&0.612	&0.347 &18.24	&0.627	&0.274 &17.51	&0.592	&0.356 &21.28	&0.732	&0.214\\
63798f5c6f &14.65	&0.518	&0.500 &14.75	&0.563	&0.418 &15.02	&0.587	&0.457 &18.39	&0.696	&0.337\\
669c36225b &12.39	&0.370	&0.530 &12.20	&0.390	&0.490 &12.70	&0.445	&0.500 &15.46	&0.543	&0.384\\
66fd66cbed &13.83	&0.536	&0.492 &13.90	&0.570	&0.423 &13.90	&0.602	&0.446 &15.85	&0.654	&0.378\\
6d22162561 &12.44	&0.352	&0.573 &12.52	&0.365	&0.529 &13.47	&0.445	&0.543 &14.42	&0.474	&0.468\\
6d81c5ab0d &15.33	&0.496	&0.459 &15.48	&0.522	&0.385 &15.29	&0.524	&0.430 &18.51	&0.613	&0.322\\
6e11e7f4fe &17.59	&0.714	&0.472 &17.93	&0.731	&0.404 &18.92	&0.779	&0.415 &18.49	&0.760	&0.377\\
70eac6ff18 &13.67	&0.398	&0.480 &13.38	&0.403	&0.423 &13.22	&0.387	&0.479 &16.16	&0.529	&0.348\\
71b2dc8a2a &17.08	&0.613	&0.454 &17.34	&0.651	&0.367 &17.50	&0.673	&0.373 &20.64	&0.762	&0.238\\
75fbbe4673 &13.76	&0.362	&0.524 &13.48	&0.369	&0.469  &13.94	&0.398	&0.504 &16.10	&0.466	&0.406\\
7705a2edd0 &14.24	&0.400	&0.559 &14.90	&0.435	&0.482 &16.32	&0.564	&0.513 &17.47	&0.576	&0.467\\
7a9f97660b &15.45	&0.473	&0.512 &15.48	&0.520	&0.425 &15.12	&0.549	&0.474 &18.36	&0.612	&0.348\\
7da3db9905 &12.92	&0.450	&0.521 &12.83	&0.455	&0.477 &13.48	&0.472	&0.494 &13.78	&0.504	&0.467\\
800cf88687 &13.75	&0.318	&0.524 &13.42	&0.321	&0.490 &13.67	&0.350	&0.518 &15.43	&0.376	&0.467\\
8324b3ca22 &12.22	&0.304	&0.592 &12.35	&0.327	&0.548 &13.04	&0.417	&0.551 &13.76	&0.410	&0.522\\
85cd0e9211 &14.24	&0.365	&0.528 &14.40	&0.386	&0.469 &14.28	&0.432	&0.498 &16.48	&0.472	&0.429\\
8b9fb9d9f1 &13.16	&0.288	&0.551 &13.26	&0.299	&0.520 &13.63	&0.342	&0.576 &15.26	&0.381	&0.499\\
8cb2e97d26 &14.88	&0.382	&0.521 &14.90	&0.404	&0.424 &14.38	&0.427	&0.494 &20.39	&0.617	&0.300\\
8fdc5130f0 &15.44	&0.496	&0.448 &15.33	&0.511	&0.387 &15.13	&0.532	&0.420 &17.94	&0.616	&0.318\\
90cb7ef953 &15.69	&0.500	&0.493 &15.44	&0.519	&0.448 &15.52	&0.551	&0.461 &15.70	&0.547	&0.463\\
917e9c8985 &14.17	&0.387	&0.477 &13.90	&0.395	&0.436 &14.27	&0.418	&0.469 &16.48	&0.529	&0.345\\

\hline
    \end{tabular}
}
    \caption{Per-scene 3DGS artifact restoration quantitative comparison on DL3DV-Res (Part 1).}
    \label{tab:dl3dv_Res_per_scene_video_restoration_part1}
\end{table}

\begin{table}[ht]
    \renewcommand{\tabcolsep}{4.5pt}
    \centering
    \resizebox{0.98\linewidth}{!}{
    \begin{tabular}{l|ccc|ccc|ccc|ccc}
    & \multicolumn{3}{c|}{Artifact} & \multicolumn{3}{c|}{Difix3D+} & \multicolumn{3}{c|}{GenFusion}& \multicolumn{3}{c}{GSFixer (Ours)}\\
    & PSNR$\uparrow$ & SSIM$\uparrow$ & LPIPS$\downarrow$ & PSNR$\uparrow$ & SSIM$\uparrow$ & LPIPS$\downarrow$ & PSNR$\uparrow$ & SSIM$\uparrow$ & LPIPS$\downarrow$ & PSNR$\uparrow$ & SSIM$\uparrow$ & LPIPS$\downarrow$ \\
\hline
91afb9910b &13.96	&0.409	&0.515 &13.86	&0.415	&0.480 &15.08	&0.483	&0.490 &16.52	&0.504	&0.437\\
946f49be73 &15.67	&0.454	&0.436 &16.25	&0.473	&0.337 &16.06	&0.484	&0.385 &18.61	&0.574	&0.287\\
9641a1ed79 &15.73	&0.523	&0.506 &16.07	&0.561	&0.432 &15.29	&0.562	&0.489 &17.48	&0.626	&0.391\\
9c8c0e0fad &10.43	&0.246	&0.697 &10.36	&0.200	&0.635 &11.78	&0.267	&0.669 &13.96	&0.262	&0.569\\
9cbc554864 &12.32	&0.256	&0.550 &12.11	&0.259	&0.529 &12.76	&0.296	&0.543 &14.95	&0.382	&0.456\\
9e9a89ae6f &15.54	&0.627	&0.412 &15.38	&0.640	&0.340 &14.37	&0.638	&0.412 &17.25	&0.703	&0.319\\
9fb0588ff0 &14.06	&0.390	&0.435 &14.09	&0.407	&0.343 &14.69	&0.401	&0.389 &18.59	&0.613	&0.265\\
a17a984ca9 &12.76	&0.338	&0.524 &12.63	&0.351	&0.489 &12.90	&0.375	&0.517 &13.24	&0.395	&0.473\\
a401469cb0 &12.77	&0.358	&0.572 &12.67	&0.371	&0.539 &13.49	&0.432	&0.564 &14.64	&0.457	&0.510\\
a62c330f54 &11.94	&0.317	&0.572 &11.81	&0.332	&0.550 &13.27	&0.414	&0.555 &13.86	&0.428	&0.528\\
a62f9a1c63 &13.93	&0.400	&0.502 &14.00	&0.412	&0.420 &13.63	&0.406	&0.494 &17.54	&0.545	&0.343\\
a726c1112a &12.30	&0.463	&0.530 &13.11	&0.510	&0.481 &14.38	&0.606	&0.481 &14.36	&0.592	&0.490\\
adb95f29c1 &14.73	&0.394	&0.508 &15.19	&0.398	&0.470 &15.82	&0.481	&0.508 &17.92	&0.504	&0.444\\
adf35184a1 &15.87	&0.526	&0.523 &15.86	&0.559	&0.438 &15.68	&0.588	&0.484 &20.09	&0.691	&0.339\\
af0d7039e6 &12.33	&0.310	&0.509 &12.01	&0.303	&0.479 &12.80	&0.338	&0.494 &13.27	&0.365	&0.479\\
b2076bc723 &11.04	&0.267	&0.593 &10.76	&0.274	&0.585 &11.83	&0.332	&0.582 &12.61	&0.400	&0.529\\
b3bf9079b4 &16.08	&0.461	&0.441 &16.23	&0.496	&0.371 &15.98	&0.502	&0.414 &18.68	&0.579	&0.326\\
b4f53094fd &13.97	&0.384	&0.487 &13.75	&0.388	&0.429 &13.60	&0.380	&0.487 &16.73	&0.506	&0.349\\
b5faa2a8ce &13.90	&0.429	&0.537 &14.29	&0.443	&0.469 &17.50	&0.566	&0.468 &16.88	&0.560	&0.434\\
b6d1134cb0 &11.97	&0.256	&0.559 &11.74	&0.262	&0.524 &12.31	&0.274	&0.562 &12.71	&0.303	&0.511\\
b92b499c9b &18.20	&0.705	&0.348 &19.01	&0.744	&0.243 &20.26	&0.787	&0.255 &22.30	&0.842	&0.171\\
ba55c875d2 &17.50	&0.612	&0.394 &17.57	&0.633	&0.312 &17.87	&0.653	&0.333 &20.59	&0.721	&0.248\\
c076929db6 &17.44	&0.608	&0.373 &18.14	&0.630	&0.291 &18.54	&0.640	&0.295 &23.28	&0.738	&0.196\\
c37109a55e &12.10	&0.247	&0.547 &11.94	&0.253	&0.516 &12.83	&0.280	&0.545 &13.39	&0.298	&0.508\\
c37726ce77 &13.50	&0.287	&0.537 &13.30	&0.289	&0.479 &14.13	&0.342	&0.508 &14.81	&0.377	&0.453\\
c455899acf &16.43	&0.539	&0.465 &16.11	&0.533	&0.417 &16.31	&0.569	&0.438 &15.94	&0.555	&0.437\\
cbd44beb04 &14.85	&0.433	&0.498 &14.64	&0.450	&0.442 &14.32	&0.455	&0.511 &18.03	&0.588	&0.341\\
cc08c0bdc3 &17.30	&0.530	&0.413 &17.85	&0.555	&0.324 &17.08	&0.545	&0.411 &20.60	&0.655	&0.290\\
cd9c981eeb &12.40	&0.300	&0.539 &12.41	&0.313	&0.492 &12.82	&0.351	&0.510 &14.82	&0.438	&0.420\\
ceb252f5d4 &12.86	&0.401	&0.529 &12.66	&0.419	&0.496 &13.04	&0.467	&0.528 &14.87	&0.521	&0.445\\
d1b3a0b37a &13.59	&0.374	&0.538 &13.36	&0.367	&0.508 &14.65	&0.404	&0.523 &13.70	&0.382	&0.531\\
d3812aad53 &10.99	&0.341	&0.576 &11.03	&0.362	&0.540 &12.07	&0.440	&0.554 &11.90	&0.408	&0.555\\
d3af8212ae &15.82	&0.425	&0.481 &16.10	&0.456	&0.405 &16.51	&0.491	&0.447 &21.05	&0.623	&0.272\\
d4fbeba016 &12.10	&0.171	&0.565 &11.74	&0.164	&0.540 &12.45	&0.199	&0.560 &12.86	&0.206	&0.556\\
d8de66037b &13.47	&0.320	&0.520 &13.12	&0.321	&0.492 &13.70	&0.376	&0.503 &14.60	&0.395	&0.456\\
d904ae2998 &11.63	&0.322	&0.568 &11.52	&0.320	&0.552 &12.94	&0.411	&0.556 &13.22	&0.399	&0.558\\
d9b6376623 &12.74	&0.315	&0.570 &12.89	&0.320	&0.535 &13.57	&0.380	&0.559 &14.71	&0.408	&0.503\\
d9f4c746e6 &15.33	&0.516	&0.409 &15.59	&0.543	&0.329 &16.54	&0.564	&0.346 &19.76	&0.698	&0.210\\
dac9796dd6 &14.38	&0.503	&0.510 &14.11	&0.535	&0.450 &13.56	&0.535	&0.472 &18.62	&0.700	&0.273\\
dafa9c7cbd &11.35	&0.137	&0.599 &11.21	&0.132	&0.579 &12.26	&0.164	&0.587 &12.58	&0.190	&0.581\\
ddfdcfdf02 &12.63	&0.293	&0.557 &12.28	&0.297	&0.514 &12.66	&0.310	&0.553 &14.55	&0.401	&0.450\\
ded5e4b46a &17.44	&0.563	&0.506 &17.23	&0.580	&0.458 &17.80	&0.628	&0.479 &18.78	&0.658	&0.425\\
df04f58064 &14.02	&0.277	&0.529 &13.91	&0.285	&0.481 &13.90	&0.286	&0.532 &17.45	&0.442	&0.376\\
df29c22586 &15.52	&0.402	&0.425 &15.60	&0.421	&0.338 &15.80	&0.427	&0.397&20.31	&0.589	&0.281\\
df4f9d9a0a &13.12	&0.473	&0.321 &13.26	&0.504	&0.430 &14.42	&0.563	&0.452 &17.58	&0.662	&0.340\\
e5684b3292 &13.90	&0.415	&0.545 &13.93	&0.447	&0.497 &13.25	&0.506	&0.537 &15.40	&0.550	&0.453\\
e78f8cebd2 &12.51	&0.206	&0.562 &12.45	&0.209	&0.534 &13.32	&0.241	&0.546 &14.71	&0.296	&0.491\\
e8ce51b6ab &14.38	&0.491	&0.499 &14.17	&0.513	&0.438 &14.27	&0.529	&0.486 &18.52	&0.687	&0.284\\
e9360e7a89 &12.13	&0.319	&0.554 &11.84	&0.319	&0.521 &12.31	&0.335	&0.565 &12.67	&0.362	&0.516\\
eb4cf52988 &13.75	&0.417	&0.543 &13.53	&0.431	&0.518 &14.12	&0.476	&0.532 &15.67	&0.526	&0.478\\
ec1e44d4dc &13.81	&0.452	&0.534 &13.50	&0.481	&0.489 &13.51	&0.506	&0.497 &16.37	&0.606	&0.353\\
ec305787b7 &13.68	&0.476	&0.484 &13.77	&0.509	&0.415 &14.16	&0.569	&0.389 &14.99	&0.575	&0.375\\
ed16328235 &14.50	&0.214	&0.549 &15.04	&0.217	&0.500 &15.89	&0.277	&0.551 &17.07	&0.293	&0.483\\
ef59aac437 &17.15	&0.331	&0.515 &17.32	&0.338	&0.443 &17.71	&0.381	&0.498 &19.97	&0.465	&0.361\\
f004c810d9 &14.48	&0.594	&0.405 &14.51	&0.624	&0.317 &14.21	&0.631	&0.343 &19.79	&0.760	&0.210 \\
f477ffc4b3 &14.75	&0.528	&0.490 &14.38	&0.551	&0.421 &13.7	&0.534	&0.461 &17.21	&0.635	&0.336\\
f71ac346cd &13.23	&0.230	&0.534 &13.06	&0.224	&0.499 &13.72	&0.254	&0.529 &15.63	&0.328	&0.440\\
f7aaea9ac6 &13.11	&0.247	&0.529 &13.14	&0.253	&0.473 &13.76	&0.260	&0.530 &16.34	&0.388	&0.416\\
fb2c0499c2 &14.93	&0.442	&0.469 &15.61	&0.481	&0.375 &16.38	&0.530	&0.381 &19.14	&0.608	&0.284\\
fb3b73f1d3 &11.78	&0.132	&0.622 &11.72	&0.125	&0.610 &12.80	&0.180	&0.630 &13.41	&0.189	&0.588\\
ff59239865 &12.35	&0.432	&0.533 &12.81	&0.420	&0.481 &14.28	&0.455	&0.493 &17.72	&0.516	&0.395\\
average  &14.12	&0.405	&0.509 &14.14	&0.419	&0.455 &14.56	&0.453	&0.486 &16.72 & 0.520 & 0.399\\
\hline
    \end{tabular}
}
    \caption{Per-scene 3DGS artifact restoration quantitative comparison on DL3DV-Res (Part 2).}
    \label{tab:dl3dv_Res_per_scene_video_restoration_part2}
\end{table}

\begin{table}[ht]
    \renewcommand{\tabcolsep}{4.5pt}
    \centering
    \begin{tabular}{l|ccc|ccc|ccc}
    & \multicolumn{3}{c|}{Difix3D+}  & \multicolumn{3}{c|}{GenFusion}& \multicolumn{3}{c}{GSFixer (Ours)}\\
    & PSNR$\uparrow$ & SSIM$\uparrow$ & LPIPS$\downarrow$ & PSNR$\uparrow$ & SSIM$\uparrow$ & LPIPS$\downarrow$ & PSNR$\uparrow$ & SSIM$\uparrow$ & LPIPS$\downarrow$ \\
\hline
0bfdd020cf  &15.96	&0.498	&0.494 &15.16	&0.511	&0.510 &17.04	&0.526	&0.476\\
2beaca3189  &14.33	&0.401	&0.423 &13.31	&0.380	&0.546 &14.78	&0.457	&0.526\\
5a69d1027b  &14.70	&0.359	&0.452 &15.49	&0.407	&0.514 &16.56	&0.435	&0.465\\
9c8c0e0fad  &11.50	&0.260	&0.605 &12.10	&0.306	&0.648 &13.77	&0.343	&0.623\\
032dee9fb0  &15.66	&0.528	&0.509 &12.74	&0.506	&0.489 &16.57	&0.565	&0.455\\
85cd0e9211  &17.78	&0.550	&0.383 &16.96	&0.580	&0.437 &18.03	&0.578	&0.441\\
91afb9910b  &14.09	&0.508	&0.518 &15.14	&0.576	&0.481 &16.20	&0.589	&0.485\\
165f5af8bf  &16.49	&0.476	&0.395 &15.63	&0.496	&0.458 &16.85	&0.516	&0.439\\
374ffd0c5f  &18.62	&0.639	&0.418 &18.72	&0.669	&0.404 &20.03	&0.692	&0.393\\
457e9a1ae7  &16.28	&0.583	&0.458 &17.13	&0.640	&0.417 &18.33	&0.660	&0.416\\
669c36225b  &12.01	&0.449	&0.576 &11.78	&0.496	&0.539 &13.28	&0.503	&0.527\\
9641a1ed79  &15.37	&0.610	&0.621 &15.19	&0.615	&0.517 &16.46	&0.612	&0.495\\
56452d9cd9  &13.76	&0.400	&0.549 &13.74	&0.418	&0.526 &14.93	&0.420	&0.513\\
493816813d  &14.91	&0.358	&0.404 &14.91	&0.414	&0.485 &16.49	&0.612	&0.496\\
0853979305  &18.31	&0.630	&0.368 &17.21	&0.653	&0.417 &18.62	&0.671	&0.394\\
adb95f29c1  &14.07	&0.427	&0.538 &13.09	&0.452	&0.565 &15.24	&0.485	&0.550\\
b3bf9079b4  &15.43	&0.555	&0.374 &15.07	&0.583	&0.394 &16.79	&0.590	&0.396\\
ba55c875d2  &17.18	&0.613	&0.386 &16.02	&0.600	&0.421 &18.10	&0.635	&0.397\\
d1b3a0b37a  &13.90	&0.443	&0.527 &13.39	&0.456	&0.523 &14.78	&0.486	&0.520\\
d904ae2998  &12.94	&0.439	&0.647 &10.92	&0.344	&0.593 &13.06	&0.416	&0.568\\
dac9796dd6  &13.90	&0.561	&0.518 &14.29	&0.559	&0.453 &15.79	&0.610	&0.443\\
dafa9c7cbd  &11.68	&0.169	&0.497 &11.81	&0.224	&0.643 &12.64	&0.250	&0.624\\
df29c22586  &19.31	&0.538	&0.238 &19.84	&0.585	&0.370 &21.13	&0.633	&0.318\\
e9360e7a89  &14.50	&0.463	&0.490 &13.52	&0.446	&0.508 &14.82	&0.492	&0.526\\
ec305787b7  &14.23	&0.587	&0.515 &13.37	&0.511	&0.435 &15.44	&0.640	&0.413\\
ed16328235  &14.83	&0.281	&0.485 &13.58	&0.306	&0.606 &15.83	&0.355	&0.573\\
f004c810d9  &17.53	&0.729	&0.314 &17.76	&0.757	&0.321 &18.85	&0.763	&0.325\\
ff59239865  &12.56	&0.416	&0.547 &12.09	&0.443	&0.583 &13.51	&0.467	&0.574\\
average  &15.07	&0.4816	&0.473 &14.64	&0.498	&0.493 &16.21	&0.536	&0.478\\

\hline
    \end{tabular}
    \caption{Comparison of per-scene 3D reconstruction results of DL3DV with 3 input views.}
    \label{tab_sup:dl3dv_per_scene_3d_recon_sparse_view3}
\end{table}
\begin{table}[ht]
    \renewcommand{\tabcolsep}{4.5pt}
    \centering
    \begin{tabular}{l|ccc|ccc|ccc}
    & \multicolumn{3}{c|}{Difix3D+}  & \multicolumn{3}{c|}{GenFusion}& \multicolumn{3}{c}{GSFixer (Ours)}\\
    & PSNR$\uparrow$ & SSIM$\uparrow$ & LPIPS$\downarrow$ & PSNR$\uparrow$ & SSIM$\uparrow$ & LPIPS$\downarrow$ & PSNR$\uparrow$ & SSIM$\uparrow$ & LPIPS$\downarrow$ \\
\hline
0bfdd020cf &17.54	&0.547	&0.394 &17.39	&0.584	&0.405 &18.21	&0.582	&0.399\\
2beaca3189 &17.56	&0.592	&0.250 &18.39	&0.638	&0.359 &18.37	&0.627	&0.361\\
5a69d1027b &17.35	&0.471	&0.322 &16.97	&0.501	&0.407 &18.47	&0.516	&0.365\\
9c8c0e0fad &15.15	&0.342	&0.425 &14.69	&0.378	&0.535 &16.14	&0.399	&0.472\\
032dee9fb0 &20.48	&0.690	&0.265 &19.65	&0.727	&0.293 &21.13	&0.722	&0.285\\
85cd0e9211 &20.60	&0.673	&0.258 &20.14	&0.692	&0.337 &22.32	&0.754	&0.266\\
91afb9910b &17.56	&0.622	&0.346 &18.53	&0.673	&0.366 &18.64	&0.672	&0.365\\
165f5af8bf &18.06	&0.551	&0.302 &17.76	&0.570	&0.374 &18.71	&0.576	&0.349\\
374ffd0c5f &23.98	&0.807	&0.195 &24.29	&0.827	&0.224 &24.50	&0.819	&0.227\\
457e9a1ae7 &19.48	&0.694	&0.347 &20.48	&0.752	&0.322 &20.93	&0.735	&0.343\\
669c36225b &15.42	&0.566	&0.400 &16.24	&0.300	&0.396 &16.64	&0.612	&0.412\\
9641a1ed79 &18.05	&0.658	&0.470 &18.17	&0.668	&0.418 &18.47	&0.651	&0.419\\
56452d9cd9 &16.35	&0.422	&0.409 &17.70	&0.511	&0.420 &18.84	&0.562	&0.373\\
493816813d &16.89	&0.482	&0.326 &16.93	&0.515	&0.406 &17.54	&0.522	&0.401\\
0853979305 &22.35	&0.771	&0.212 &22.39	&0.807	&0.245 &22.71	&0.789	&0.260\\
adb95f29c1 &16.78	&0.516	&0.435 &15.73	&0.526	&0.482 &18.87	&0.583	&0.383\\
b3bf9079b4 &18.84	&0.662	&0.253 &18.29	&0.696	&0.296 &18.56	&0.676	&0.301\\
ba55c875d2 &20.78	&0.742	&0.227 &20.18	&0.764	&0.273 &21.37	&0.751	&0.265\\
d1b3a0b37a &16.22	&0.516	&0.435 &16.82	&0.566	&0.426 &16.66	&0.538	&0.450\\
d904ae2998 &14.36	&0.480	&0.541 &13.10	&0.446	&0.533 &13.97	&0.480	&0.526\\
dac9796dd6 &19.83	&0.735	&0.285 &20.63	&0.767	&0.278 &20.87	&0.768	&0.279\\
dafa9c7cbd &13.32	&0.226	&0.418 &13.29	&0.270	&0.573 &14.21	&0.279	&0.541\\
df29c22586 &24.24	&0.713	&0.150 &24.64	&0.737	&0.256 &24.57	&0.724	&0.216\\
e9360e7a89 &15.80	&0.540	&0.372 &15.63	&0.559	&0.426 &16.09	&0.569	&0.430\\
ec305787b7 &18.53	&0.701	&0.310 &19.27	&0.733	&0.299 &19.01	&0.723	&0.309\\
ed16328235 &16.18	&0.358	&0.382 &16.29	&0.398	&0.504 &16.50	&0.400	&0.502\\
f004c810d9 &22.77	&0.859	&0.157 &22.94	&0.872	&0.202 &23.62	&0.874	&0.194\\
ff59239865 &16.89	&0.549	&0.339 &17.59	&0.611	&0.403 &19.05	&0.600	&0.378\\
average   &18.26	&0.5895	&0.329 &18.36	&0.610	&0.374 &19.11	&0.625	&0.360\\

\hline
    \end{tabular}
    \caption{Comparison of per-scene 3D reconstruction results of DL3DV with 6 input views.}
    \label{tab_sup:dl3dv_per_scene_3d_recon_sparse_view6}
\end{table}
\begin{table}[ht]
    \renewcommand{\tabcolsep}{4.5pt}
    \centering
    \begin{tabular}{l|ccc|ccc|ccc}
    & \multicolumn{3}{c|}{Difix3D+}  & \multicolumn{3}{c|}{GenFusion}& \multicolumn{3}{c}{GSFixer (Ours)}\\
    & PSNR$\uparrow$ & SSIM$\uparrow$ & LPIPS$\downarrow$ & PSNR$\uparrow$ & SSIM$\uparrow$ & LPIPS$\downarrow$ & PSNR$\uparrow$ & SSIM$\uparrow$ & LPIPS$\downarrow$ \\
\hline
0bfdd020cf &19.15	&0.582	&0.320  &19.70	&0.635	&0.366 &20.03	&0.605	&0.352\\
2beaca3189 &19.79	&0.694	&0.185  &20.52	&0.725	&0.284 &20.20	&0.706	&0.286\\
5a69d1027b &18.69	&0.542	&0.270  &18.80	&0.578	&0.356 &19.83	&0.581	&0.316\\
9c8c0e0fad &16.76	&0.411	&0.380  &16.67	&0.442	&0.482 &17.35	&0.442	&0.436\\
032dee9fb0 &22.93	&0.787	&0.169  &22.73	&0.829	&0.211 &22.89	&0.801	&0.209\\
85cd0e9211 &22.88	&0.762	&0.174  &22.60	&0.778	&0.267 &22.48	&0.757	&0.262\\
91afb9910b &20.45	&0.715	&0.248  &20.28	&0.737	&0.301 &20.51	&0.733	&0.295\\
165f5af8bf &19.41	&0.597	&0.255  &19.12	&0.617	&0.330 &19.67	&0.611	&0.316\\
374ffd0c5f &26.67	&0.871	&0.120  &27.01	&0.886	&0.163 &26.67	&0.874	&0.160\\
457e9a1ae7 &21.66	&0.761	&0.262  &22.05	&0.800	&0.281 &22.21	&0.777	&0.299\\
669c36225b &17.03	&0.636	&0.324  &17.91	&0.690	&0.336 &17.60	&0.656	&0.362\\
9641a1ed79 &18.85	&0.675	&0.419  &19.30	&0.702	&0.389 &19.32	&0.676	&0.389\\
56452d9cd9 &17.32	&0.473	&0.393  &19.01	&0.594	&0.370 &19.29	&0.561	&0.372\\
493816813d &18.35	&0.563	&0.249  &18.11	&0.581	&0.338 &18.88	&0.596	&0.324\\
0853979305 &24.54	&0.835	&0.152  &24.66	&0.856	&0.192 &24.86	&0.841	&0.199\\
adb95f29c1 &18.29	&0.557	&0.343  &18.17	&0.598	&0.399 &18.88	&0.584	&0.383\\
b3bf9079b4 &19.37	&0.717	&0.212  &19.76	&0.762	&0.247 &19.87	&0.737	&0.254\\
ba55c875d2 &22.98	&0.808	&0.170  &22.79	&0.825	&0.207 &23.79	&0.818	&0.200\\
d1b3a0b37a &18.41	&0.607	&0.315  &18.97	&0.660	&0.332 &19.06	&0.623	&0.355\\
d904ae2998 &16.80	&0.548	&0.408  &16.03	&0.548	&0.443 &16.71	&0.553	&0.430\\
dac9796dd6 &22.06	&0.807	&0.207  &22.00	&0.815	&0.232 &22.42	&0.812	&0.231\\
dafa9c7cbd &14.33	&0.276	&0.385  &13.92	&0.297	&0.533 &14.80	&0.309	&0.512\\
df29c22586 &25.66	&0.759	&0.115  &26.14	&0.782	&0.228 &26.01	&0.774	&0.172\\
e9360e7a89 &17.95	&0.626	&0.289  &18.66	&0.668	&0.339 &18.15	&0.643	&0.352\\
ec305787b7 &21.07	&0.776	&0.203  &21.65	&0.806	&0.220 &21.54	&0.790	&0.224\\
ed16328235 &18.44	&0.460	&0.295  &17.98	&0.477	&0.417 &18.18	&0.467	&0.414\\
f004c810d9 &25.11	&0.899	&0.126  &25.50	&0.911	&0.157 &25.40	&0.906	&0.157\\
ff59239865 &18.37	&0.613	&0.272  &18.88	&0.658	&0.369 &20.13	&0.654	&0.333\\
average   &20.12	&0.656	&0.259  &20.32	&0.688	&0.314 &20.60	&0.675	&0.307\\

\hline
    \end{tabular}
    \caption{Comparison of per-scene 3D reconstruction results of DL3DV with 9 input views.}
    \label{tab_sup:dl3dv_per_scene_3d_recon_sparse_view9}
\end{table}
\begin{table}[ht]
    \renewcommand{\tabcolsep}{4.5pt}
    \centering
    \begin{tabular}{l|ccc|ccc|ccc}
    & \multicolumn{3}{c|}{Difix3D+}  & \multicolumn{3}{c|}{GenFusion}& \multicolumn{3}{c}{GSFixer (Ours)}\\
    & PSNR$\uparrow$ & SSIM$\uparrow$ & LPIPS$\downarrow$ & PSNR$\uparrow$ & SSIM$\uparrow$ & LPIPS$\downarrow$ & PSNR$\uparrow$ & SSIM$\uparrow$ & LPIPS$\downarrow$ \\
\hline
\multicolumn{10}{c}{3 Views} \\
\hline
bicycle  &13.02  &0.177   &0.632  & 14.75 & 0.247 & 0.632 &15.80 &0.284 &0.613 \\
bonsai  &13.86  &0.397  &0.549  & 13.87 & 0.408 & 0.549 &13.42 &0.434 &0.554\\
counter  &13.96 &0.394  &0.519  & 15.20 & 0.470 & 0.519 &15.25 &0.472 &0.505\\
flowers & 11.81 &0.154  &0.693 & 12.86 & 0.202 & 0.693 &13.80 &0.288 &0.654\\
garden  & 14.34 &0.258  &0.573 & 16.13 & 0.289 & 0.573 &17.42 &0.326 &0.537\\
kitchen  & 15.56 &0.371  &0.537 & 16.14 & 0.419 & 0.537 &16.61 &0.424 &0.503\\
room  & 14.15 &0.465  &0.427 & 16.23 & 0.568 & 0.427 &15.54 &0.540 &0.441\\
stump  & 15.35 &0.215 &0.620 & 16.51 & 0.296 & 0.620 &17.62 &0.320 &0.603\\
treehill & 13.19 &0.253  &0.649 & 13.57 & 0.311 & 0.649 &15.03 &0.319 &0.619\\
average  &13.92  &0.298  &0.578  & 15.03 & 0.357  & 0.578 &15.61 &0.370 &0.559\\
\hline
\multicolumn{10}{c}{6 Views} \\
\hline
bicycle  &15.58 &0.253 &0.594 & 16.16 & 0.292 & 0.570 &17.32 &0.302 &0.549\\
bonsai  &16.30 &0.525 &0.427  & 17.08 & 0.545 & 0.426 &17.29 &0.556 &0.440\\
counter  &16.09 &0.490 &0.413  & 17.06 & 0.545 & 0.426 &17.12 &0.530 &0.425\\
flowers  &12.66 &0.178 &0.572  & 13.69 & 0.224& 0.632 &14.10 &0.216 &0.594\\
garden &17.58 &0.360 &0.363  & 18.62 & 0.396 & 0.452 &18.98 &0.400 &0.429\\
kitchen  &17.95 &0.534 &0.310  & 18.62 & 0.556 & 0.388 &18.78 &0.555 &0.371\\
room  &15.07 &0.525 &0.460  & 17.73 & 0.628 & 0.387 &16.76 &0.606 &0.403\\
stump  &17.51 &0.284 &0.485  & 17.77 & 0.321 & 0.570 &18.65 &0.328 &0.543\\
treehill &14.70 &0.293 &0.588  & 15.36 & 0.353 & 0.585 &16.46 &0.341 &0.546\\
average  &15.94 &0.382  &0.468   & 16.90 & 0.430 & 0.494 &17.27 &0.426 &0.478\\
\hline
\multicolumn{10}{c}{9 Views} \\
\hline
bicycle  &16.99 &0.288 &0.498  & 16.89 & 0.313 & 0.541 &17.89 &0.322 &0.511\\
bonsai  &18.54 &0.661 &0.343 & 19.43 & 0.661 & 0.343 &19.53 &0.654 &0.356\\
counter  &17.42 &0.552 &0.352  & 18.18 & 0.607 & 0.368 &18.82 &0.600 &0.362\\
flowers  &13.97 &0.218 &0.499  & 14.50 & 0.253& 0.593 &14.98 &0.244 &0.534\\
garden  &19.18 &0.453 &0.282  & 19.79 & 0.471 & 0.393 &20.10 &0.473 &0.367\\
kitchen  &20.08 &0.610 &0.243  & 20.56 & 0.635 & 0.316 &20.58 &0.623 &0.314\\
room  &17.96 &0.626 &0.343  & 19.72 & 0.700 & 0.317 &18.96 &0.666 &0.333\\
stump  &18.50 &0.340 &0.425  & 19.09 & 0.376 & 0.522 &19.59 &0.378 &0.496\\
treehill &15.25 &0.322 &0.537  & 16.42 & 0.387 & 0.566 &17.24 &0.366 &0.507\\
average  &17.54  &0.452  &0.391   & 18.29 & 0.489  & 0.507 &18.63 &0.481 &0.420\\
\hline
    \end{tabular}
    \caption{Per-scene 3D sparse view reconstruction quantitative comparison on Mip-NeRF 360.}
    \label{tab_sup:mipnerf_per_scene_3d_recon}
\end{table}

\end{document}